\documentclass{article}

\usepackage[accepted]{icml2024}

\usepackage[utf8]{inputenc} % allow utf-8 input
\usepackage[T1]{fontenc}    % use 8-bit T1 fonts
\usepackage{hyperref}       % hyperlinks
\usepackage{url}            % simple URL typesetting
\usepackage{booktabs}       % professional-quality tables
\usepackage{amsfonts}       % blackboard math symbols
\usepackage{nicefrac}       % compact symbols for 1/2, etc.
\usepackage{microtype}      % microtypography
\usepackage{natbib}

\newcommand{\miss}{\textsc{LASER}\xspace}

\usepackage[font=small,labelfont=bf]{caption}
\usepackage{wrapfig}

\usepackage[utf8]{inputenc} % allow utf-8 input
\usepackage[T1]{fontenc}    % use 8-bit T1 fonts
\usepackage{url}            % simple URL typesetting
\usepackage{booktabs}       % professional-quality tables
\usepackage{amsfonts}       % blackboard math symbols
\usepackage{nicefrac}       % compact symbols for 1/2, etc.
\usepackage{microtype}      % microtypography
\usepackage{tabu}
\usepackage{multicol}
\usepackage{soul}
\usepackage{bbm}
\usepackage{lipsum}
\usepackage{kantlipsum}
\usepackage{tabularx}

\usepackage{cancel}

\usepackage{amsmath,amssymb,amsfonts,amsthm, dsfont, color}
\makeatletter
\g@addto@macro\normalsize{%
  \setlength\abovedisplayskip{7pt}
  \setlength\belowdisplayskip{7pt}
  \setlength\abovedisplayshortskip{7pt}
  \setlength\belowdisplayshortskip{7pt}
}
\makeatother
\usepackage{algorithm}
\usepackage{algorithmicx}
 \usepackage[noend]{algpseudocode}
\usepackage{mathtools}
\usepackage{graphicx}
\usepackage{textcomp}
\usepackage{enumitem}
\usepackage{float}
\usepackage{nicefrac}

\usepackage{wrapfig}
\usepackage{mathtools}
\usepackage{cuted}
\definecolor{MyGreen1}{RGB}{20,180,40}
\usepackage{multirow, tikz,float}

\usepackage{nicefrac,color,mathrsfs,float}
\usepackage{multirow,caption,tikz}
\usetikzlibrary{shapes.misc, positioning}
\usetikzlibrary{decorations.pathreplacing}
\usetikzlibrary{arrows.meta, shapes,patterns}

\tikzset{
  block/.style    = {draw, thick, rectangle, minimum width = 2em},
sblock/.style      = {draw, thick, rectangle, minimum height = 2em,
minimum width = 2em}, 
}

\newcommand{\Expect}{\mathbb{E}}

\newcommand{\norm}[1]{\|#1\|}
\newcommand{\normbig}[1]{\left \|#1 \right \|}

\renewcommand{\tilde}{\widetilde}

\newcommand{\argmin}{\mathop{\operatorname{argmin}}}

\newcommand{\calC}{\mathcal{C}}
\newcommand{\calD}{\mathcal{D}}

\usepackage{comment}

\newcommand{\snr}{\mathrm{SNR}}	
\newcommand{\awgn}{\mathcal{Z}}
\newcommand{\awgnp}{\mathcal{Z}_P}

\newcommand{\gsgd}{\textsc{Z-SGD}\xspace}

\newcommand{\signum}{\textsc{Signum}\xspace}
\newcommand{\randk}{\textsc{Random-K}\xspace}
\newcommand{\sketch}{\textsc{Sketching}\xspace}
\newcommand{\adsgd}{\textsc{A-DSGD}\xspace}

\newcommand{\alg}{\mathrm{ALG}}
\newcommand{\bigo}{\mathcal{O}}

\newcommand{\setx}{\{ \bx_i \}}
\newcommand{\setg}{\{ \bg_i \}}

\newcommand{\mnist}{\textsc{Mnist}\xspace}
\newcommand{\cifarten}{\textsc{Cifar10}\xspace}
\newcommand{\cifarhun}{\textsc{Cifar100}\xspace}
\newcommand{\resnet}{\textsc{ResNet18}\xspace}
\newcommand{\wikileaks}{\textsc{WikiText-103}\xspace}

\newcommand{\gpt}{\textsc{GPT}\xspace}
\newcommand{\gptwo}{\textsc{GPT-2}\xspace}

\newcommand{\missinflu}{\lambda_\miss}

\newcommand{\iterates}[1]{\{#1\}_{t \geq 0}}

\newcommand{\fungap}[1]{f({\btheta}_{#1}) - f_\ast}

\newcommand{\fullgt}{\nabla f(\btheta_t)}

\providecommand{\rankcx}{\mathcal{C}_r}

\algnewcommand{\Hyperparameters}[1]{%
  \State \textbf{hyperparameters:} #1
}
\algnewcommand{\Input}[1]{%
  \State \textbf{input:}
  \Statex \hspace*{\algorithmicindent}\parbox[t]{.8\linewidth}{\raggedright #1}
}
\algnewcommand{\Initialization}[1]{%
  \State \textbf{initialization:}
  \Statex \hspace*{\algorithmicindent}\parbox[t]{.8\linewidth}{\raggedright #1}
}
\algnewcommand{\Prerequisites}[1]{%
  \State \textbf{prerequisites:}
  \Statex \hspace*{\algorithmicindent}\parbox[t]{.8\linewidth}{\raggedright #1}
}
\algnewcommand{\Initialize}[1]{%
  \State \textbf{initialize} #1
}
\algnewcommand{\Notation}[1]{%
  \State \textbf{notation:} #1
}
\algnewcommand{\Note}[1]{%
  \State \textbf{note:} #1
}
\algblock[At]{At}{EndAt}
\algblockdefx[At]{At}{EndAt}[1]{\textbf{at} #1 \textbf{do}}{\textbf{end at}}
\algblock[For]{For}{EndFor}
\algblockdefx[For]{For}{EndFor}[1]{\textbf{for} #1 \textbf{do}}{\textbf{end for}}
\algblock[Prerequisites]{Prerequisites}{EndPrerequisites}
\algblockdefx[Prerequisites]{Prerequisites}{EndPrerequisites}[0]{\textbf{prerequisites}}{}

\usepackage{thmtools}

\newtheorem{theorem}{Theorem}
\newtheorem{lemma}{Lemma}
\newtheorem{assumption}{Assumption}

\newtheorem{remark}{Remark}

\newtheorem{definition}{Definition}

\usepackage{prettyref,xspace}
\usepackage{tikz}

\newrefformat{cond}{Condition~\ref{#1}}
\newrefformat{eq}{Eq.~\eqref{#1}}
\newrefformat{thm}{Thm~\ref{#1}}
\newrefformat{th}{Thm~\ref{#1}}
\newrefformat{chap}{Chapter~\ref{#1}}
\newrefformat{sec}{Sec.~\ref{#1}}
\newrefformat{algo}{Alg.~\ref{#1}}
\newrefformat{fig}{Fig.~\ref{#1}}
\newrefformat{tab}{Table~\ref{#1}}
\newrefformat{rmk}{Remark~\ref{#1}}
\newrefformat{clm}{Claim~\ref{#1}}
\newrefformat{def}{Definition~\ref{#1}}
\newrefformat{cor}{Corollary~\ref{#1}}
\newrefformat{lmm}{Lemma~\ref{#1}}
\newrefformat{prop}{Proposition~\ref{#1}}
\newrefformat{pr}{Proposition~\ref{#1}}
\newrefformat{app}{App.~\ref{#1}}
\newrefformat{prob}{Problem~\ref{#1}}
\newrefformat{ques}{Question~\ref{#1}}
\newrefformat{notee}{Note~\ref{#1}}
\newrefformat{assump}{Assumption~\ref{#1}}
\newrefformat{issuee}{Issue ~\ref{#1}}
\newrefformat{fix}{Fix ~\ref{#1}}

\newcommand{\reals}{\mathbb{R}}
\newcommand{\naturals}{\mathbb{N}}

\newcommand{\inner}[2]{\langle {#1}, {#2}  \rangle}

\newcommand{\calN}{\mathcal{N}}
\newcommand{\calO}{\mathcal{O}}
\newcommand{\vect}[1]{\boldsymbol{#1}}

\newcommand{\ba}{\vect{a}}
\newcommand{\bb}{\vect{b}}

\newcommand{\be}{\vect{e}}

\newcommand{\bg}{\vect{g}}

\newcommand{\bp}{\vect{p}}
\newcommand{\bq}{\vect{q}}

\newcommand{\bx}{\vect{x}}
\newcommand{\by}{\vect{y}}

\newcommand{\bM}{\vect{M}}

\newcommand{\bP}{\vect{P}}
\newcommand{\bQ}{\vect{Q}}

\newcommand{\bY}{\vect{Y}}
\newcommand{\bZ}{\vect{Z}}

\newcommand{\btheta}{\vect{\theta}}

\newcommand{\balpha}{\vect{\alpha}}
\newcommand{\bbeta}{\vect{\beta}}
\newcommand{\bxi}{\vect{\xi}}

\newcommand{\matrx}[2]{\reals^{#1 \times #2}}

\newcommand{\define}{\triangleq}

\newcommand{\pth}[1]{\left( #1 \right)}
\newcommand{\qth}[1]{\left[ #1 \right]}
\newcommand{\sth}[1]{\left\{ #1 \right\}}

\newcommand{\ie}{i.e.\xspace}

\newcommand{\id}{\mathbf{I}}
\mathchardef\mhyphen="2D

\definecolor{MyGreen1}{RGB}{20,180,40}

\usepackage{siunitx}

\newcommand{\0}{\mathbf{0}}

\let\ast\star %MJ: ast often used for convex conjugates, star for optima

\icmltitlerunning{\miss: Linear Compression in Wireless Distributed Optimization}

\begin{document}

\twocolumn[
\icmltitle{\miss: Linear Compression in Wireless Distributed Optimization}

\icmlsetsymbol{equal}{*}

\begin{icmlauthorlist}
\icmlauthor{Ashok Vardhan Makkuva}{equal,epfl}
\icmlauthor{Marco Bondaschi}{equal,epfl}
\icmlauthor{Thijs Vogels}{epfl}
\icmlauthor{Martin Jaggi}{epfl}
\icmlauthor{Hyeji Kim}{uta}
\icmlauthor{Michael Gastpar}{epfl}
\end{icmlauthorlist}

\icmlaffiliation{epfl}{School of Computer and Communication Sciences, EPFL, Lausanne, Switzerland}
\icmlaffiliation{uta}{Department of Electrical and Computer Engineering, UT Austin, Austin, TX, USA}

\icmlcorrespondingauthor{Ashok Vardhan Makkuva}{ashok.makkuva@epfl.ch}

\icmlkeywords{Distributed Leaning, Low-rank compression}

\vskip 0.3in
]

\printAffiliationsAndNotice{\icmlEqualContribution}

% !TEX root = main.tex

\begin{abstract}
Data-parallel SGD is the de facto algorithm for distributed optimization, especially for large scale machine learning. Despite its merits, communication bottleneck is one of its persistent issues. Most compression schemes to alleviate this either assume noiseless communication links, or fail to achieve good performance on practical tasks. In this paper, we close this gap and introduce \textsc{LASER}: {\bf L}ine{\bf A}r Compre{\bf S}sion in Wir{\bf E}less Dist{\bf R}ibuted Optimization. \textsc{LASER} capitalizes on the inherent low-rank structure of gradients and transmits them efficiently over the noisy channels. Whilst enjoying theoretical guarantees similar to those of the classical SGD, \textsc{LASER} shows consistent gains over baselines on a variety of practical benchmarks. In particular, it outperforms the state-of-the-art compression schemes on challenging computer vision and GPT language modeling tasks. On the latter, we obtain $50$-$64 \%$ improvement in perplexity over our baselines for noisy channels. Code is available at \url{https://github.com/Bond1995/LASER}.
\looseness=-1
 % With \textsc{GPT-2} on \textsc{WikiText-103}

%MJ: i removed the dataset names below here as too specific (and mnist not considered challenging. also wikitext doesn't tell yet if its a language model, there are other tasks on it. so LLM remains key to mention. 
%\textsc{Mnist}, \textsc{Cifar10}, \textsc{Cifar100}, and \textsc{WikiText-103} datasets.
\end{abstract} 

% provide an information-theoretic interpretation for the superiority of our curriculum and demonstrate that without good curricula, neural network-based decoders fail to achieve the desired reliability.

% !TEX root = main.tex

\section{Introduction}
\label{sec:intro}

Distributed optimization is one of the most widely used frameworks for training large scale deep learning models \citep{bottou2018optimization, dean2012large, tang2020communicationsurvey}. In particular, data-parallel SGD is the workhorse algorithm for this task. Underpinning this approach is the \emph{communication} of large gradient vectors between the workers and the central server which performs their \emph{aggregation}. While these methods harness the inherent parallelism to reduce the overall training time, their communication cost is a major bottleneck that limits scalability to large models. Design of communication-efficient distributed algorithms is thus a must for reaping the full benefits of distributed optimization \citep{xu2020compressedsurvey}.  

% communication cost remains a major bottleneck for these methods.

% deep learning models. Mention Federated learning. Exploiting the parallelism, efficient computational resources at each nodes, this ensures faster convergence blah blah. With more nodes, larger models, and practical bandwidth constraints, communication between workers becomes a key bottleneck in training these models. Hence an important question in this arena is to design scalable and communication-efficient algorithms.

Existing approaches to reduce the communication cost can be broadly classified into two themes: (i) compressing the gradients before transmission; or (ii) utilizing the communication link for native `over-the-air' aggregation (averaging) across workers. Along (i), a number of gradient compression schemes have been designed such as quantization \citep{bernstein2018signsgd,vargaftik2022eden}, sparsification \citep{aji2017sparse, isik2022sparse}, hybrid methods \citep{jiang2018sketchml, basu2019qsparse}, and low-rank compression \citep{wang2018atomo, vogels2019powersgd}. These methods show gains over the full-precision SGD in various settings (\citet{xu2020compressedsurvey} is a detailed survey). Notwithstanding the merits, their key shortcoming is that they assume a \emph{noiseless} communication link between the clients and the server. In settings such as federated learning with differential privacy or wireless communication, these links are noisy. Making them noiseless requires error-correcting codes which exacerbates the latency, as the server needs to wait till it receives the gradient from each worker before aggregating \citep{guo2020analog}.

% A common approach to address the network bandwidth issue is to compress the transmitted gradients and hence make better use of network resources. Over the past few decades, a number of successful gradient compression schemes have been designed. Such as quantization, sparsification, and hybrid schemes, etc. Underpinning these approaches is the assumption that the communication link is noiseless and that central server exactly receives the transmitted gradients from each client, which is usually achieved with help of a forward error correction code. However the number of radio resource blocks (?) required scales linearly with the number of clients under this approach and hence increases latency. 

Under theme (ii), communication cost is reduced by harnessing the physical layer aspects of (noisy) communication. In particular, the superposition nature of wireless channels is exploited to perform over-the-air averaging of gradients across workers, which reduces the latency, see e.g.~\cite{shi2020communication} and the references therein. Notable works include \adsgd \cite{amiri2020machine}, analog-gradient-aggregation \cite{guo2020analog, zhu2019broadband}, channel aware quantization \cite{chang2020mac}, etc. However, to the best of our knowledge, the majority of these approaches are restricted to synthetic datasets and shallow neural networks (often single layer) and do not scale well to the practical neural network models (which we verify in \prettyref{sec:main_results}). This leads to a natural question:

\emph{Can we design efficient and practical gradient compression schemes for noisy communication channels?}

\begin{table*}
  \centering
  \small
  \begin{minipage}[t]{0.52\textwidth}
  % vspace{-1mm}
    \centering
    \includegraphics[width=\textwidth]{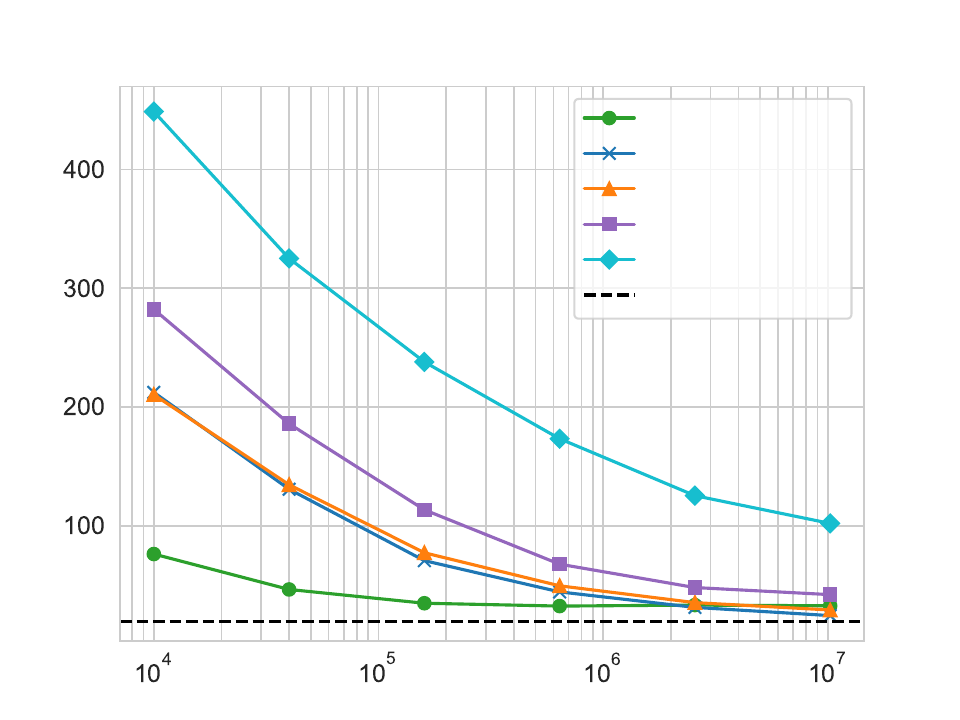}
      \put(-145,-2){\fontsize{8}{3}\selectfont Power budget}
      \put(-250,87){\rotatebox[origin=t]{90}{\fontsize{8}{3}\selectfont Perplexity}}
      \put(-80,157){\fontsize{6}{3}\selectfont LASER}
      \put(-80,148){\fontsize{6}{3}\selectfont Z-SGD}
      \put(-80,139){\fontsize{6}{3}\selectfont Sketching}
      \put(-80,130){\fontsize{6}{3}\selectfont Random-K}    
      \put(-80,121){\fontsize{6}{3}\selectfont Signum}
      \put(-80,112){\fontsize{6}{3}\selectfont Noiseless SGD}
      \vspace{3pt}
    \captionof{figure}{Final test perplexity after 20k iterations \emph{(lower is better)} vs. power budget for \gptwo language modeling on \textsc{WikiText-103}. \miss consistently requires orders-of-magnitude less power than other methods for the same perplexity.}
    \label{fig:llm}
  \end{minipage}
  \hspace{12pt}
  \begin{minipage}[b]{0.42\textwidth}
    \caption{Power required \emph{(lower is better)} to reach the target perplexity on \textsc{WikiText-103}. \ref{eq:awgn_sgd} sends the uncompressed gradients directly, while \miss sends a rank-4 approximation. \miss requires $16 \times $ less power than \textsc{Z-SGD} to achieve the target perplexity over a wide interval. In the very-high-power regime with perplexity close to that of the noiseless \ref{eq:sgd}, we see no power gains.}
    \vspace{10pt}
	\label{tab:llm-power-ratio}
	\begin{tabularx}{\textwidth}{c c c c}
    \toprule
    \multirow{2}{*}{\bf Target} &
    \multicolumn{2}{c}{\bf Power required} & 
    \multirow{2}{*}{\bf Reduction} \\
    
    \cmidrule(lr){2-3}
    
      & \textsc{Z-SGD}  & \miss & \\
    
    \cmidrule(lr){1-1}\cmidrule(lr){2-3}\cmidrule(lr){4-4}
    
    $80$ &  $160\,\mathrm{K}$  & $10\,\mathrm{K}$ & $16\times$\\
    $50$ & $640\,\mathrm{K}$ & $40\,\mathrm{K}$  & $16\times$\\
    $40$ & $2560\,\mathrm{K}$ & $160\,\mathrm{K}$  & $16\times$\\
    $35$ & $2560\,\mathrm{K}$ & $160\,\mathrm{K}$ & $16\times$\\
    \bottomrule
    \end{tabularx}
  \end{minipage}
  \vspace{-5mm}
\end{table*}

In this work, we precisely address this and propose \miss, a principled gradient compression scheme for distributed training over wireless noisy channels. Specifically, we make the following contributions:
\begin{itemize}
    \item Capitalizing on the inherent low-rank structure of the gradients, \miss efficiently computes these low-rank factors and transmits them reliably over the noisy channel while allowing the gradients to be averaged in transit (\prettyref{sec:powercomm}). 
    \item We show that \miss enjoys similar convergence rate as that of the classical SGD for both quasi-convex and non-convex functions, except for a small additive constant depending on the channel degradation (\prettyref{thm:main}).
    \item 
    We empirically demonstrate the superiority of \miss over the baselines on the challenging tasks of (i) language modeling with \gptwo $\rightarrow$ \wikileaks and (ii) image classification with \resnet $\rightarrow$ (\cifarten, \cifarhun) and \textsc{1-layer NN} $\rightarrow$ \mnist. With high gradient compression ($165 \times $), \miss achieves $50$-$64\%$ perplexity improvement in the low and moderate power regimes on \wikileaks. To the best of our knowledge, \miss is the first to exhibit such gains for GPT language modeling (\prettyref{sec:main_results}). 

\end{itemize}

% setup. theory. results. 

% Main contributions 
% \begin{itemize}
%     \item Framework: distributed training over wireless channels - gradients are sent over noisy channels 
%     \item Key insight: Leverage the low-rank property of gradients. 
%     \item Theoretical result: Single-user over wireless channels - derived optimal power scaling and convergence bound (rank-1) 
%     \item Inexact rank components (step 1) 
%     \item Error feedback - Tx does not know the precise error any more (step 2)
%     \item non-convex: numerical results (step 3) 
% \end{itemize}

% Why analog transmission of gradients? 
% keep the compression linear -- (0) raw gradients, (i) compressed sensing (linear matrix times the gradient), (ii) power sgd 
% Assumptions: what if the transmissions are not synchronized? y1,y2,...yk. There, a naive method would be to estimate g1,...,gk (either via MMSE or gi=yi) and then sum them up? 
% Power scaling? 

{\bf Notation.} Euclidean vectors and matrices are denoted by bold letters $\bx, \by, \bM$, etc. $\| \cdot \| $ denotes the Frobenius norm for matrices and the $\ell_2$-norm for Euclidean vectors. $\mathcal{O}(\cdot)$ is an upper bound subsuming universal constants whereas $\tilde{\mathcal{O}}(\cdot)$ hides any logarithmic problem-variable dependencies.

\section{Background}
\label{sec:background}

{\bf Distributed optimization}. Consider the (synchronous) data-parallel distributed setting where we minimize an objective $f:\reals^d \to \reals $ defined as the empirical loss on a global dataset $\calD=\{(\bx_j, y_j)\}_{j=1}^N $: 
\looseness=-1
\begin{align*}
\min_{\btheta \in \reals^d} f(\btheta), \quad f(\btheta) \define \frac{1}{N} \sum_{j=1}^N \ell(\bx_j, y_j ;\btheta),
\end{align*}
where $\ell(\cdot)$ evaluates the loss for each data sample $(\bx_j, y_j)$ on model $\btheta$. In this setup, there are $k$ (data-homogeneous) training clients, where the $i^{\text{th}}$ client has access to a stochastic gradient oracle $\bg_i$, e.g. mini-batch gradient on a set of samples randomly chosen from $\calD$, such that $\Expect[\bg_i|\btheta] = \nabla f(\btheta)$ for all $\btheta \in \reals^d$. 
%In this setup, there are $k$ training nodes/workers/clients that have each access to a stochastic gradient oracle $\bg$ such that $\Expect[\bg|\btheta] = \nabla f(\btheta)$ for all $\btheta \in \reals^d$. For instance, $\bg$ could be the mini-batch gradient on a set of samples randomly chosen from $\calD$. 
In distributed SGD \citep{robbins1951stochastic, bottou2018optimization}, the server aggregates all $\bg_i$s and performs the following updates:
\begin{equation}
\label{eq:sgd}
\tag{\textsc{SGD}}
\begin{gathered}
   \btheta_{t+1} = \btheta_t -  \gamma_t \cdot \frac{1}{k} \sum_{i=1}^k \bg_i^{(t)}, \\
   \Expect[\bg_i^{(t)}|\btheta_t] = \nabla f(\btheta_t), \quad t \geq 0,
  \end{gathered}
\end{equation}
where $\{\gamma_t \}_{t \geq 0}$ is a stepsize schedule. Implicit here is the assumption that the communication link between the clients and the server is noiseless, which we expound upon next.

% Underpinning the above SGD updates is the assumption that the central server receives the local gradients exactly, which maynot always hold as discussed in \prettyref{sec:intro}. We now introduce a canonical communication framework betwen the clients and the server below. 

% samples is split across $k$ distributed workers/clients with $\calD = \cup_{i \in [k]} \calD_i$. The local dataset $\calD_i =\{(\bx_j, y_j)\}_{j=1}^{n_i} $ is of size $n_i = N/k$ (for simplicity). The objective here is to minimize a function $f: \reals^d \to \reals$ in a distributed fashion as follows:

% where $f(\cdot)$ is defined as the average of the local loss functions $\ell_i(\cdot)$ over all the $k$ clients and $\ell_i$ is given by the empirical loss on the local dataset $\calD_i $ for a suitable loss function $\ell(\cdot)$. 

% We assume that there is a global central server that houses the parameter $\btheta$ and that all the clients share the same $\btheta$. Here we focus on the stochastic gradient setup where each worker has access to a stochastic gradient oracle $\bg_i$ of $f$ such that $\Expect[\bg_i] = \nabla f$. At every training iteration, all the clients communicate their local (stochastic) gradients $\{\bg_i\}$ to the central server to update the model parameters $\btheta$. \nb{} illustrates this setup. An important aspect of this distributed framework is the communication channel betwen the clients and the server which we explain below.
% which comprise some of the most practical and well-studied channels,
{\bf Communication model.} For the communication uplink from the clients to the server, we consider the standard wireless channel for over-the-air distributed learning \citep{amiri2020federated, guo2020analog, zhu2019broadband, chang2020mac, wei2022federated}: the \emph{additive slow-fading channel}, e.g., the classical multiple-access-channel \citep{nazer2007computation}.  The defining property of this family is the superposition of incoming wireless signals (enabling over-the-air computation) possibly corrupted together with an independent channel noise \citep{shi2020communication}. 
Specifically, we denote the channel as a (random) mapping $\awgn_P(\cdot)$ that transforms the set of (time-varying) messages transmitted by the clients $\{\bx_i \}_{ i \in [k]} \subset \reals^d $ to its noisy version 
$\by \in \reals^d$ received by the server:
\begin{align}
\begin{split}
    \by &= \awgn_P(\setx) \define \sum_{i=1}^k \bx_i + \bZ,\\
    &\hspace{5em}\norm{\bx_i}^2 \leq P_t, ~ \frac{1}{T}\sum_{t=0}^{T-1} P_t \leq P,
\end{split} \label{eq:normal_awgn}
\end{align}
% where the power $P >0 $ is fixed and known to everyone.

where the noise $\bZ \in \reals^d$ is independent of the channel inputs and has zero mean and unit variance per dimension, \ie $\Expect \norm{\bZ}^2=d$. The power constraint on each client $\norm{\bx_i}^2 \leq P_t$ at time $t$ serves as a communication cost (and budget), while the power policy $\{P_t\}$ allots the total budget $P$ over $T$ epochs as per the average power constraint \citep{wei2022federatedconvergence,amiri2020machine}.
% Do we need to talk about SNR here? 
    A key metric that captures the channel degradation quality is the signal-to-noise ratio per coordinate ($\snr$), defined as the ratio between the average signal energy ($P$) and that of the noise ($d$), \ie $\snr \define P/d$. % (for signals of size $m\times n$, $\snr \define P/mn$).
    The larger it is the better the signal fidelity. %; $\snr = \infty$ corresponds to the ideal scenario when there is no noise. 
    The power budget $P$ encourages the compression of signals: if each client can transmit the same information $\bx_i$ via fewer entries (smaller $d$), they can utilize more power per entry (higher $\snr$) and hence a more faithful signal. 
    % Given a power constraint $P$, we can achieve a higher $\snr$ by reducing $d$.

The downlink communication from the server to the clients is usually modeled as a standard broadcast channel \cite{Cover:1972}: for input ${\bx}$ with $\norm{\bx}^2 \leq P_b$, the output $\by_i = \bx + \bZ_i$, one for each of the clients. Usually in practice, $P_b \gg P$ and therefore we set $P_b = \infty$, though our results readily extend to finite $P_b$. 

% $ controlled by the server and generally subject to a power constraint $\norm{\bx}^2 \leq P_d.$ There are multiple noisy outputs 
    
%For the downlink communication from the server to the clients (broadcast channel), we assume that it is noiseless and thus the clients exactly receive what the server transmits \citep{fedoriginal17, fedoptim16, fedefficiency16}.
In the rest of the paper by channel we mean the uplink channel. The channel model in \prettyref{eq:normal_awgn} readily generalizes to the fast fading setup as discussed in \prettyref{sec:main_results}.

{\bf Gradient transmission over the channel.} In the distributed optimization setting the goal is to communicate the (time-varying) local gradients $\bg_i \in \reals^d$ to the central server over the noisy channel in \prettyref{eq:normal_awgn}. Here we set the messages $\bx_i$ as linear scaling of gradients (as we want to estimate the gradient average), \ie $\bx_i = a_i \, \bg_i$ with the scalars $a_i \in \reals$ enforcing the power constraints:
\begin{align}
    \label{eq:weightedgrad_awgn}
    \by  =  \sum_{i=1}^k a_i \, \bg_i + \bZ, \quad \norm{a_i \, \bg_i}^2 \leq P_t.
\end{align}
Now the received signal is a weighted sum of the gradients corrupted by noise, whereas we need the sum of the gradients $\sum_i \bg_i$ (upto zero mean additive noise) for the model training. Towards this goal, a common mild technical assumption is that the gradient norms $\{ \norm{\bg_i} \}$ are known at the receiver at each communication round
\citep{chang2020mac, guo2020analog} (can be relaxed in practice, \prettyref{sec:main_results}). The optimal scalars are then given by $a_i  =\sqrt{P_t}/(\max_j \norm{\bg_j}), \forall i \in [K]$, which are uniform across all the clients (\S~\ref{app:channel_transf}). 
% In other words, during each communication round the clients transmit their gradient norms to the server which then computes their maximum and broadcasts it back to them to compute $a_i$. 
Now substituting this $a_i$ in \prettyref{eq:weightedgrad_awgn} and rearranging, the effective channel can be written as
\begin{align}
    \by = \tilde{\awgn}_{P}(\setg) \define \frac{1}{k} \sum_{i=1}^k \bg_i + \frac{\max_i \norm{\bg_i}}{k \sqrt{P_t}} \, \bZ.
    \tag{noisy channel}
    \label{eq:awgn_final}
\end{align}
Equivalently, we can assume this as the actual channel model where the server receives the gradient average corrupted by a zero mean noise proportional to the gradients. Note that the noise magnitude decays in time as gradients converge to zero.
% When there is just a single worker, $k=1$, this simplifies to
% \begin{align}
%     \label{eq:awgn_single}
%     \by = \awgn_{P, \mathrm{eff}}(\bg_1) = \bg_1 + \frac{\norm{\bg_1}}{\sqrt{P}} \, \bZ.
% \end{align}
We denote $\tilde{\awgn}_{P}(\cdot)$ as simply $\awgn_P(\cdot)$ henceforth as these two mappings are equivalent.

%%%%
% Removed for brevity
%%%%
% Hence the scalars $a_i$ have to be chosen properly so as to meet the following two objecives: (i) enforcing the power constraint $P$ on gradients, and (ii) ensuring an unbiased estimate of the gradient sum at the receiver. 

{\bf \textsc{Z-SGD}.} Recall that the \ref{eq:sgd} aggregates the uncompressed gradients directly. In the presence of the \ref{eq:awgn_final}, it naturally modifies to
\begin{align}
\btheta_{t+1} = \btheta_t - \gamma_t \, \awgnp(\{\bg_i^{(t)} \} ).
    \tag{\textsc{Z-SGD}}
    \label{eq:awgn_sgd}
\end{align}
Thus \gsgd is a canonical baseline to compare against. It has two sources of stochasticity: one stemming for the stochastic gradients and the other from the channel noise. While the gradient in the \gsgd update still has the same conditional mean as the noiseless case  (zero mean Gaussian in \ref{eq:awgn_final}), it has higher variance due to the Gaussian term. When $P = \infty$, \ref{eq:awgn_sgd} reduces to \ref{eq:sgd}.

% For training the model parameters $\btheta$ at the client using the (synchronous) distributed SGD algorithm, we follow the following communication protocol at each iteration $t$: (i) all the clients compute the local gradients $\bg_i$ based on $\btheta_t$ and transmit their norms $\norm{\bg_i}$ to the client; (ii) the server transmits them back the maximum of these norms $\max_i \norm{\bg_i}$; (iii) the clients transmit $\bx_i = \sqrt{P}/(\max_j \norm{\bg_j}) \cdot \bg_i $ and the server receives $\by = \sum_i \sqrt{P}/(\max_j \norm{\bg_j}) \cdot \bg_i + \calN(0,I_d) $; (iv) the server does a simple rescaling to obtain $\by$ as in \prettyref{eq:awgn_final}: $\by = \awgn(\setg) = \sum_i \bg_i + \max_j \norm{\bg_j}/\sqrt{P} \cdot \calN(0,I_d) $; (v) serves uses $\by$ to update the parameters to $\btheta_{t+1}$ which it broadcasts to all the clients. These steps (i)-(v) are again repeated at $t+1$ and so forth. We term this algorithm {\it AWGN-SGD}.

% !TEX root = main.tex

\section{\miss: Novel linear compression cum transmission scheme} 
\label{sec:powercomm}

% !TEX root = main.tex

% \pr{NOTE for future: $\gamma$ placement, $\lambda$ placement, and redo experiments.}
In this section we describe our main contribution, \miss, a novel method to compress gradients and transmit them efficiently over noisy channels.
The central idea underpinning our approach is that, given the channel power constraint 
in \prettyref{eq:normal_awgn}, we can get a more faithful gradient signal at the receiver by transmitting its `appropriate' compressed version (fewer entries sent and hence more power per entry) as opposed to sending the full-gradient naively as in \gsgd. This raises a natural question: {\it what's a good compression scheme that facilitates this?} To address this, we posit that we can capitalize on the inherent low-rank structure of the gradient matrices \citep{martin2021implicit,mazumder2010spectral, yoshida2017spectral} for efficient gradient compression and transmission. Indeed, as illustrated below and in \prettyref{thm:main}, we can get a variance reduction of the order of the smaller dimension when the gradient matrices are approximately low-rank.
 %MJ is this obvious? do wanna say the power cost is high?

% Ashok commented. Not so.
% to utilize the fact that the bulk of the gradients transmitted during training correspond to that of the weight matrices of the neural network layers 
% (ii) to further capitalize on their inherent low-rank structure \nb{} to efficiently transmit them over the noisy AWGN channel. 
% % 
% In other words, for the same power constraint $P$, we can get a more faithful gradient signal at the receiver by transmitting its low rank components (fewer entries sent and hence more power per entry) as opposed to sending the full-gradient naively as in AWGN-SGD. 

% , where entries of $\bZ$ are sampled \iid from $\calN(0,1)$.
More concretely, let us consider the single worker case where the goal is to transmit the stochastic gradient $\bg \in \matrx{m}{m}$ (viewed as a matrix) to the server with constant power $P_t = P$. Further let's suppose that $\bg$ is approximately rank-one, \ie $\bg \approx \bp \bq^\top$, with the factors $\bp, \bq \in \reals^m $ known. If we transmit $\bg$ uncompressed over the noisy channel, as in \gsgd, the server receives $\by_{\gsgd} = \bg + (\norm{\bg}/\sqrt{P})~\bZ \in \matrx{m}{m}$. On the other hand, if we capitalize on the low-rank structure of $\bg$ and instead transmit the factors $\bp$ and $\bq$ with power $P/2$ each, the server would receive: 
\looseness=-1
\begin{equation*}
\begin{gathered}
    \by_{\bp} = \bp + (\sqrt{2} \norm{\bp}/\sqrt{P}) ~\bZ_{\bp} \in \reals^m, \\
    \by_{\bq} = \bq + (\sqrt{2} \norm{\bq}/\sqrt{P})~ \bZ_{\bq} \in \reals^m,
\end{gathered}
\end{equation*}
where $\bZ_{\bp}$ and $\bZ_{\bq}$ are the channel noise. Now we reconstruct the stochastic gradient as
\begin{multline}
        \by_{\miss} \define \by_{\bp} \by_{\bq}^\top = (\bp + (\sqrt{2} \norm{\bp}/\sqrt{P}) ~\bZ_{\bp}) \\
        \cdot (\bq + (\sqrt{2} \norm{\bq}/\sqrt{P})~ \bZ_{\bq} )^\top.
        \label{eq:miss_motiv}
\end{multline}
Conditioned on the gradient $\bg$, while the received signal $\by$ has the same mean $\bg$ under both \gsgd and \miss, we observe that for \gsgd it has variance %MJ: don't say SGD has variance. the transmitted (decoded) thing y has variance (and mention why this matters. because this term will determine the convergence speed of SGD as it's directly proportional to it). make sure the motivation paragraph here is understandable for dummies :)
$\Expect \norm{\by_{\gsgd} - \bg }^2 = \norm{\bg}^2/\snr $ with $\snr \define P/m^2$, whereas that of \miss is roughly $\norm{\bg}^2 \cdot (4/m\snr)(1+1/(m \snr))$, as further elaborated in Definition~\ref{def:cif}. 
When $\snr$ is of constant order $\Omega(1)$, we observe that the variance for \miss is roughly $\bigo(m)$ times smaller than that of \gsgd, which is significant given that variance directly affects the convergence speed of stochastic-gradient based methods \citep{bottou2018optimization}.

More generally, even if the gradients are not inherently low-rank and we only know their rank factors approximately, with standard techniques like error-feedback \citep{seide20141} we can naturally generalize the aforementioned procedure, which is the basis for \miss. 
\prettyref{algo:miss_comm} below details \miss and \prettyref{thm:main} establishes its theoretical justification. While LASER works with any power policy $\{P_t\}$ in \ref{eq:awgn_final}, it suffices to consider the constant law $P_t=P$ as justified in \prettyref{sec:power-policy}.

\subsection{Algorithm}
\label{sec:algo}

\begin{algorithm}
\caption{\miss}
\label{algo:miss_comm}
\begin{algorithmic}[1]
    \State {\bf input}: initial model parameters $\btheta \in \matrx{m}{n}$, learning rate $\gamma$, compression rank $r$, power budget $P$
    %{\color{gray}($= 0.9$ by default)}}

    \State {\bf output}: trained parameters $\btheta$
    
    % \Initialize{model parameters $\btheta \in \matrx{m}{n}$}
    
    \At{each worker $i=1,\ldots, k$}
    
    \Initialize{memory $\be_i \leftarrow \0 \in \matrx{m}{n}$}
    
    \For{each iterate $t=0, \ldots $}
    
    \State Compute a stochastic gradient $\bg_i \in \matrx{m}{n}$
    
    \State \makebox[12mm][l]{$\bM_i       $} $\leftarrow  \be_i + \gamma \bg_i $
            %\Comment{Updated gradient via error-feedback}
            
    \State \makebox[12mm][l]{$\bP_i, \bQ_i  $}   $\leftarrow \rankcx(\bM_i)$
         %\Comment{Rank-$r$ compression \citep{vogels2019powersgd}} 
         
    \State \makebox[12mm][l]{$\be_i $} $\leftarrow \bM_i - \textsc{decompress}(\rankcx(\bM_i))$ 
            %\Comment{Memorize local errors}

    \State \makebox[12mm][l]{$\balpha, \bbeta $}   $ \leftarrow \textsc{poweralloc}( \{ \rankcx(\bM_j) , \bM_j \}) $ 
            %\Comment{Power allocation}
    \State \makebox[12mm][l]{$\bY_{\bp}, ~ \bY_{\bq} $} $\leftarrow \awgn_{\balpha}(\{\bP_j \}) , ~ \awgn_{\bbeta}(\{\bQ_j \}) $
            %\Comment{Channel transmission}
            
    \State \makebox[12mm][l]{$\bg  $} $\leftarrow \textsc{decompress}(\bY_{\bp}, \bY_{\bq})$
            %\Comment{Reconstruct the gradient in $ \matrx{m}{n}$}
            
    % \State \makebox[10mm][l]{$\mm            $} $\leftarrow \lambda + \Delta^\prime$
    %     %  \Comment{Update momentum}
    \State \makebox[12mm][l]{$\btheta            $} $\leftarrow \btheta - \bg$
        %  \Comment{Update the model}
    \EndFor
    \EndAt
    \end{algorithmic}
\end{algorithm}

For distributed training of neural network models, we apply \prettyref{algo:miss_comm} to each layer independently. Further we use it only for the weight matrices (fully connected layers) and the convolutional filters (after reshaping the multi-dimensional tensors to matrices), and transmit the bias vectors uncompressed. Now we delineate the two main components of \miss: (i) Gradient compression + Error-feedback (EF), and (ii) Power allocation +  Channel transmission.

{\bf Gradient compression and error feedback (7-9).} Since we transmit low-rank gradient approximations, we use error feedback (EF) to incorporate the previous errors into the current gradient update. This ensures convergence of SGD with biased compressed gradients \citep{karimireddy2019error}. For the rank-$r$ compression of the updated gradient $\bM$, $\rankcx(\bM)$, we use the PowerSGD algorithm from \citet{vogels2019powersgd}, a linear compression scheme to compute the left and right singular components $\bP \in \matrx{m}{r}$ and $\bQ \in \matrx{n}{r}$ respectively. PowerSGD uses a single step of the subspace iteration \citep{stewart1975methods} with a warm start from the previous updates to compute these factors. The approximation error, $\bM - \bP \bQ^\top$, is then used to update the error-feedback for next iteration. Note that the clients do not have access to the channel output and only include the local compression errors into their feedback. The decompression function in line $9$ is given by $\textsc{decompress}(\bP, \bQ) \define \bP \bQ^\top \in \matrx{m}{n}$. 

{\bf Power allocation and channel transmission (10-11).} This block is similar to \prettyref{eq:miss_motiv} we saw earlier but generalized to multiple workers and higher rank. For each client, to transmit the rank-$r$ factors $\bP$ and $\bQ$ over the noisy channel, we compute the corresponding power-allocation vectors $\balpha, \bbeta \in \reals_+^r$, given by $\balpha, \bbeta = \textsc{poweralloc}(\bP,\bQ, \bM )$. This allocation is uniform across all the clients. Given these power scalars, all the clients synchronously transmit the corresponding left factors over the channel which results in $\bY_{\bp} \in \reals^{m \times r}$. Similarly for $\bY_{\bq} \in \matrx{n}{r}$. Finally, the stochastic gradient for the model update is reconstructed as $\bg = \bY_{\bp} \bY_{\bq}^\top $. For brevity we defer the full details to \S~\ref{app:channel_transf}.

 % but the main idea here is to first determine the power allocation for each rank component proportional to its singular value, $P_s \propto  \sigma_s P$ for $s \in [r]$. Then we further divide $(P_s)_{s \in [r]}$ into $\balpha$ and $\bbeta$ according to \prettyref{lmm:power_rankr} and average it across users.

% !TEX root = main.tex

\subsection{Theoretical results}
\label{sec:theory}

We now provide theoretical justification for \miss for learning parameters in $\matrx{m}{n}$ with $m \leq n$ (without loss of generality). While our algorithm works for any number of clients, for the theory we consider $k=1$ to illustrate the primary gains with our approach. Our results readily extend to the multiple clients setting following \citet{cordonnier2018convex}. Specifically, \prettyref{thm:main} below highlights that the asymptotic convergence rate of \miss is \emph{almost the same as that of the classical \ref{eq:sgd}}, except for a small additive constant $\lambda_{\miss} $ which is $\bigo(m)$ times smaller than that of \gsgd. Our results hold for both quasi-convex and arbitrary non-convex functions. We start with the preliminaries.

\begin{definition}[\bf Channel influence factor]\label{def:cif}
For any compression cum transmission algorithm $\mathrm{ALG}$, let $\by_{\alg}(\bg)$ be the reconstructed gradient at the server after transmitting $\bg$ over the noisy channel. Then the channel influence factor $\lambda_{\alg}$ is defined as
\begin{align}
    \lambda_{\alg} \define \frac{\Expect_{\bZ}\norm{\by_{\alg}(\bg) - \bg}^2}{\norm{\bg}^2}.
    \label{eq:chanel_infl}
\end{align}
\end{definition}
The influence factor gauges the effect of the channel on the variance of the final gradient $\by_\alg$: if the original stochastic gradient $\bg$ has variance $\sigma^2$ with respect to the actual gradient $\nabla f$, then $\by_\alg$ has $ (1+\lambda_\alg) \sigma^2$. Note that this variance directly affects the convergence speed of the SGD and hence the smaller $\lambda_\alg$ is, the better the compression scheme is. %Note that this variance directly affects the convergence speed of the SGD and hence the smaller $\lambda_\alg$ is better the compression scheme.
In view of this, the following fact (\S~\ref{app:channel_influ}) illustrates the crucial gains of \miss compared to \gsgd, which are roughly of order $\bigo(m)$:
\begin{align}
    \lambda_\miss &\leq  \frac{4}{(m/r) \snr} \pth{1+ \frac{1}{(n/r) \snr}} \notag\\
    &\ll \frac{1}{\snr} = \lambda_{\gsgd}.
    \label{eq:lambda_better}
\end{align}
In the low-rank \citep{vogels2019powersgd} and constant-order SNR regime where $r = \bigo(1)$ and $\snr = \Omega(1)$, we observe that $\lambda_\miss$ is roughly $\bigo(m)$ times smaller than $\lambda_{\gsgd}$. In other words, the effective $\snr$ seen by \miss roughly gets boosted to $\bigo(m \, \snr)$ due to capitalizing on the low-rank factors whereas \gsgd perceives only the standard factor $\snr$. Constant-order SNR, \ie $P/mn = \Omega(1)$, means that the energy used to transmit each coordinate is roughly a constant, analogous to the constant-order bits used in quantization schemes \citep{vargaftik2021drive}. In fact, a weaker condition that $P/4r^2 > 1$
 suffices (\S~\ref{app:constant_snr}). With a slight abuse of notation, we denote the first upper bounding quantity in \prettyref{eq:lambda_better} as $\lambda_\miss$ too and $\textsc{Decompress}(\calC_r(\cdot))$ as $\calC_r(\cdot)$ for brevity.

We briefly recall the standard assumptions for SGD convergence following the framework in \citet{bottou2018optimization} and \citet{ stich2019error}. 

\begin{assumption}
\label{assump:quasi}
The objective $f:\reals^{m \times n} \to \reals$ is differentiable and $\mu$-quasi-convex for a constant $\mu \geq 0$ with respect to $\btheta_{\ast}$, \ie
$
        f(\btheta) - f(\btheta_\ast) + \frac{\mu}{2}\norm{\btheta - \btheta_\ast}^2 \leq \inner{\nabla f(\btheta)}{\btheta - \btheta_\ast}, \, \, \forall \,\btheta \in \matrx{m}{n}.        
$
\end{assumption}

\begin{assumption}
\label{assump:smooth}
f is $L$-smooth for some $L >0$, \ie
    $
         f(\btheta') \leq f(\btheta) + \langle \nabla f(\btheta), \btheta' - \btheta \rangle + \frac{L}{2} \norm{\btheta'- \btheta}^2, \, \, \forall \, \btheta, \btheta' \in \reals^{m \times n}.
    $
\end{assumption}

\begin{assumption}
\label{assump:grad_oracle}
For any $\btheta$, a gradient oracle $\bg(\btheta, \bxi) = \nabla f(\btheta) + \bxi$, and conditionally independent noise $\bxi$, there exist scalars $(M, \sigma^2) \geq 0$ such that
    $
      \Expect \qth{\bxi| \btheta} = 0,  \, \, \Expect [\norm{\bxi}^2| \btheta ] \leq M \norm{\nabla f(\btheta)}^2 + \sigma^2.
    $
\end{assumption}

\begin{assumption}
\label{assump:compressor}
The compressor $\rankcx(\cdot)$ satisifes the $\delta_r$-\textit{compression} property: there exists a $\delta_r \in [0,1]$ such that
    $
      \Expect_{\rankcx}\norm{\rankcx(\bM) - \bM}^2 \leq (1-\delta_r) \norm{\bM} ^2, \, \, \forall \, \bM \in \matrx{m}{n}.
    $
\end{assumption}

$\delta_r$-compression is a standard assumption in the convergence analysis of Error Feedback SGD (\textsc{EF-SGD}) \citep{stich2020error}. It ensures that the norm of the feedback memory remains bounded. We make the following assumption on the influence factor $\lambda_\miss$, which ensures that the overall composition of the channel and compressor mappings, $\awgnp(\rankcx(\cdot))$, still behaves nicely. 
\begin{assumption}
\label{assump:influence}
The channel influence factor $\lambda_\miss$ satisfies
    $
        \lambda_\miss \leq 1/(10(2/\delta_r + M)).
    $
\end{assumption}

We note that a similar assumption is needed for convergence even in the hypothetical ideal scenario when the clients have access to the channel output (\S~\ref{app:channel_influ}), which we do not have. This bound can be roughly interpreted as $\lambda_\miss = \mathcal{O}(\delta_r)$. We are now ready to state our main result.

\begin{theorem}[\bf \miss  convergence]
% \leavevmode
\label{thm:main}
Let $\{\btheta_t\}_{t \geq 0}$ be the \miss iterates (Alg.~\ref{algo:miss_comm}) with constant stepsize schedule $\{ \gamma_t = \gamma \}_{t \geq 0}$ and suppose Assumptions~\ref{assump:smooth}-\ref{assump:influence} hold. Denote $\btheta_\ast \define \argmin_{\btheta} f(\btheta), f_\ast \define f(\btheta_\ast)$, and $\tau \define 10 L \pth{\frac{2}{\delta_r} + M}$. Then for $k=1$, 
% align=left, widest=i, leftmargin=*, itemsep=0pt, topsep=0pt, partopsep=0pt
\begin{enumerate}[label={\upshape(\roman*)}, nosep,leftmargin=12pt,itemsep=2pt]
    \item if $f$ is $\mu$-quasi convex for $\mu >0$, there exists a stepsize $\gamma \leq \frac{1}{\tau (1+\lambda_\miss)}$ such that
        \begin{align*}
        \Expect &f(\btheta_{\mathrm{out}}) - f_{\ast} =\\
        &\tilde{\mathcal{O}} \Bigg(\tau (1+\lambda_\miss) \norm{\btheta_0 - \btheta^\ast}^2\exp \pth{\frac{-\mu T}{\tau  (1+\lambda_\miss) } } \\
        &\hspace{12em}+\frac{\sigma^2 (1+\lambda_\miss)}{\mu T} \Bigg),
    \end{align*}
    where $\btheta_{\mathrm{out}}$ is chosen from $\{ \btheta \}_{t=0}^{T-1}$ such that $\btheta_{\mathrm{out}} = \btheta_t $ with probability $(1- \mu \gamma/2)^{-t}$.
    \item  if $f$ is $\mu$-quasi convex for $\mu = 0$, there exists a stepsize $\gamma \leq \frac{1}{\tau (1+\lambda_\miss)}$ such that
        \begin{multline*}
        \Expect f(\btheta_{\mathrm{out}}) - f_{\ast} =\mathcal{O} \Bigg(\frac{\tau  \norm{\btheta_0 - \btheta^\ast}^2 (1+\lambda_\miss) }{T}\\
        + \sigma \norm{\btheta - \btheta_\ast} 
        \sqrt{\frac{1+\lambda_\miss}{T}}\Bigg) ,
    \end{multline*}
    where $\btheta_{\mathrm{out}}$ is chosen uniformly at random from $\{ \btheta \}_{t=0}^{T-1}$.
    \item  if $f$ is an arbitrary non-convex function, there exists a stepsize $\gamma \leq \frac{1}{\tau (1+\lambda_\miss)}$ such that
        \begin{multline*}
        \Expect \norm{\nabla f(\btheta_{\mathrm{out}})}^2 = \mathcal{O} \Bigg( \frac{\tau \norm{f(\btheta_0) - f_\ast}^2 (1+\lambda_\miss) }{T}\\ 
        + \sigma \sqrt{ \frac{L(f(\btheta) - f_\ast) (1+\lambda_\miss)}{T} } \Bigg),
    \end{multline*}
    where $\btheta_{\mathrm{out}}$ is chosen uniformly at random from $\{ \btheta \}_{t=0}^{T-1}$.
    \item \ref{eq:awgn_sgd} obeys the convergence bounds (i)-(iii) with $\delta_r =1$ and $\lambda_\miss$ replaced by $\lambda_{\gsgd}$.
\end{enumerate}

\end{theorem}

{\bf \miss vs. \gsgd.} Thus the asymptotic rate of \miss is dictated by the timescale $(1+\lambda_\miss)/T$, very close to the $1/T$ rate for the classical \ref{eq:sgd}. In contrast, \gsgd has the factor $(1+ \lambda_{\gsgd})/T$ with $ \lambda_{\gsgd} = \mathcal{O}(m) \, \lambda_\miss$.

{\bf Multiple clients.} As all the workers in LASER (Alg. \ref{algo:miss_comm}) apply the same linear operations for gradient compression (via PowerSGD), \prettyref{thm:main} can be extended to (homogenous) multiple workers by shrinking the constants $\sigma^2, \snr, \lambda_\miss$, and $\lambda_{\gsgd}$ by a factor of $k$, following \citet{cordonnier2018convex}.%etc.
% We capitalize on the convergence analysis tools from \cite{stich2019error} for EF-SGD. 
\begin{proof}
(Sketch) First we write the \miss iterates $\{\btheta_t \}_{t \geq 0}$ succinctly as 
\begin{align*}
\btheta_{t+1} &= \btheta_t - \awgn(\rankcx(\be_t + \gamma_t \bg_t)), \\
\be_{t+1} &= (\be_t + \gamma_t \bg_t) - \rankcx(\be_t + \gamma_t \bg_t).
\end{align*}
% Now we define a sequence of virtual iterates $\{\tilde{\btheta_t }\}_{t \geq 0}$ given by $\tilde{\btheta_t} = \btheta_t - \be_t$. 
First we establish a bound on the gap to the optimum, $\Expect \norm{{\btheta}_{t+1} - \btheta_\ast}^2 $, by the descent lemma (\prettyref{lmm:onestep_miss}). This optimality gap depends on the behavior of the error updates via $\Expect \norm{\be_t}^2$, which we characterize by the error-control lemma (\prettyref{lmm:misserror_control}). When $f$ is quasi-convex, these two lemmas help us establish a recursive inequality between the optimality gap $\Expect f(\btheta_{t+1}) - f_\ast$ at time $t+1$ and with that of at time $t$: $\Expect f(\btheta_{t}) - f_\ast$. Upon unrolling this recursion and taking a weighted summation, \prettyref{lmm:quasi_recursion} establishes the desired result. In the case of non-convexity, the same idea helps us to control $\Expect \norm{\nabla f(\btheta_t)}^2$ in a similar fashion and when combined with \prettyref{lmm:noncvx_recursion}, yields the final result. The proof for \gsgd is similar. 
\looseness = -1
\end{proof}

\section{Experimental results} 
\label{sec:main_results} 

We empirically demonstrate the superiority of \miss over state-of-the-art baselines on a variety of benchmarks, summarized in \prettyref{tab:tasks}.

%
%In this section, we present empirical results comparing our method to a diverse set of baselines and over different tasks. Our aim is to prove the practicality of LASER, which is shown to outperform current state-of-the-art algorithms in gradient compression and gradient communication over noisy channels. Furthermore, this improvement is shown to be consistent, regardless of the difficulty of the tasks and the size of the models considered.
%
%Our study aims at shading light on the interplay between the different parameters in our model, namely, power constraint, test accuracy, and compression factor. We show that LASER achieves state-of-the-art accuracy in any power regime, beating the other algorithms when running at the same compression factor.

\begin{table}[t]
\centering
% \small
\caption{Benchmarks for evaluating \miss. Baseline refers to the  noiseless \ref{eq:sgd}.}
    \small
\label{tab:tasks}
\begin{tabularx}{\linewidth}{l l l l}
    \toprule
    {\bf Model} &
    {\bf Dataset} &
    {\bf Metric} &
    {\bf Baseline} \\
    
    \cmidrule(lr){1-1}\cmidrule(lr){2-2}\cmidrule(lr){3-4}

    \textsc{GPT-2} ($123.6 \, \mathrm{M} $) &
    \textsc{WikiText} &
    Perplexity &
    $19.2$ \\
    
    \cmidrule(lr){1-4}

    \multirow{2}{*}{\resnet ($11.2 \, \mathrm{M} $)}&
    \cifarten & \multirow{3}{*}{\shortstack{Top-1\\ accuracy}}
    &
    $93.0\%$ \\
    & \cifarhun &
     &
    $ 73.1\% $ \\

    \cmidrule(lr){1-1}\cmidrule(lr){2-2}\cmidrule(lr){4-4}
    
    $1$-\textsc{layer NN} ($7850$) & \mnist &   & $92.3\%$ \\

    \bottomrule
\end{tabularx}
\end{table}

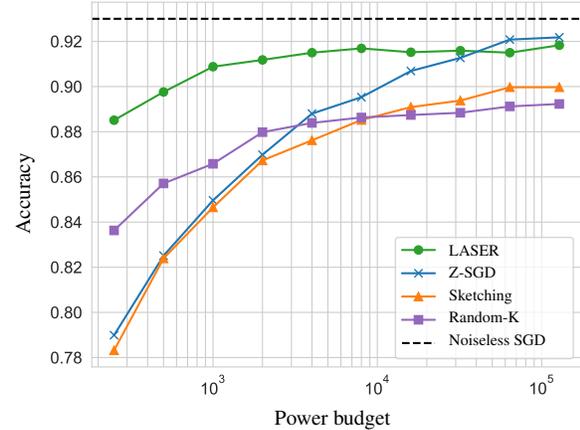
\begin{figure}
    \vspace{-5mm}
    \centering
    \includegraphics[width=0.49\textwidth]{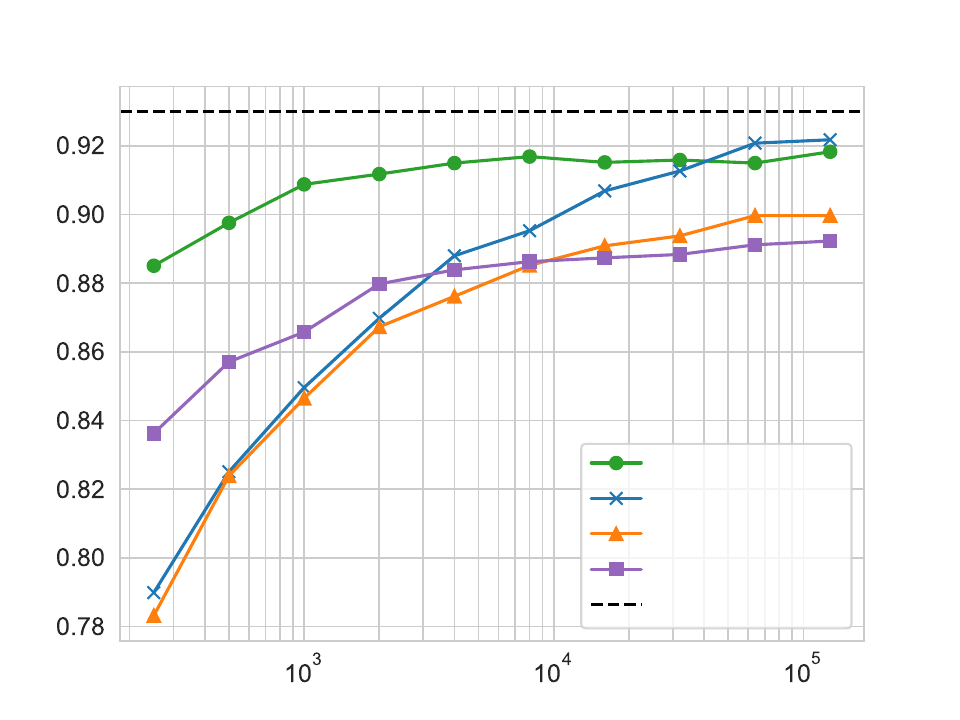}
      \put(-140,-2){\fontsize{8}{3}\selectfont Power budget}
      \put(-238,80){\rotatebox[origin=t]{90}{\fontsize{8}{3}\selectfont Accuracy}}
      \put(-74,62){\fontsize{6}{3}\selectfont LASER}
      \put(-74,53.5){\fontsize{6}{3}\selectfont Z-SGD}
      \put(-74,45){\fontsize{6}{3}\selectfont Sketching}
      \put(-74,36.5){\fontsize{6}{3}\selectfont Random-K}    
      \put(-74,28){\fontsize{6}{3}\selectfont Noiseless SGD}
      \vspace{3pt}
    \captionof{figure}{Test accuracy (\emph{higher the better}) for a given power budget on \textsc{Cifar-10} for different algorithms. \miss demonstrates consistent accuracy gains over the baselines over a wide range of power levels.}
    \label{fig:cifar10}
    \vspace{-5mm}
\end{figure}

\begin{table}
    \caption{Power required (\emph{lower the better}) to reach the given target accuracy on \textsc{Cifar-10}. \miss requires $16 \times $ lesser power than the \textsc{Z-SGD} to achieve the same targetaccuracy. Equivalently, \miss tolerates more channel noise than the \textsc{Z-SGD} for the same target accuracy as is partly supported by our theoretical analysis. }
	\label{tab:power-ratio}
    \centering
	\begin{tabularx}{0.38\textwidth}{c c c c}
    \toprule
    \multirow{2}{*}{\bf Target} &
    \multicolumn{2}{c}{\bf Power required} & 
    \multirow{2}{*}{\bf Reduction} \\
    
    \cmidrule(lr){2-3}
    
      & \miss & \textsc{Z-SGD} & \\
    
    \cmidrule(lr){1-1}\cmidrule(lr){2-3}\cmidrule(lr){4-4}
    
    $88\%$ & $250$ & $4000$ & $16\times$\\
    $89\%$ & $500$ & $8000$ & $16\times$\\
    $90\%$ & $1000$ & $16000$ & $16\times$\\
    $91\%$ & $2000$ & $32000$ & $16\times$\\
    \bottomrule
    \end{tabularx}
\end{table}

{\bf Setup.} We consider four challenging tasks of practical interest: (i) \gpt language modeling on \wikileaks, and (ii, iii, iv) image classification on \mnist, \cifarten and \cifarhun. For the language modeling, we use the \gptwo like architecture following \citet{pagliardini-llm} (\S~\ref{app:add_exp}). \resnet is used for the \textsc{Cifar} datasets. For \mnist, we use a $1$-hidden-layer network for a fair comparison with \citet{amiri2020machine}. For distributed training of these models, we consider $k=4$ clients for language modeling and $k=16$ for image classification. We simulate the noisy channel by sampling $\bZ \sim \calN(0, \id_d)$. To gauge the performance of algorithms over a wide range of noisy conditions, we vary the power $P$ geometrically in the range $[0.1, 10]$ for \mnist, $[250, 128000]$ for \cifarten and \cifarhun, and $[10000, 1024 \times 10000]$ for \wikileaks. The chosen ranges can be roughly split into low-moderate-high power regimes. Recall from \ref{eq:awgn_final} that the smaller the power, the higher the noise in the channel.

{\bf Baselines.} We benchmark \miss against three different sets of baselines: (i) \ref{eq:awgn_sgd}, (ii) \signum, \randk, \sketch, and (iii) \adsgd. \gsgd sends the uncompressed gradients directly over the noisy channel and acts as a canonical baseline. The algorithms in (ii) are state-of-the-art distributed compression schemes for noiseless communication \citep{vogels2019powersgd}. \signum \citep{bernstein2018signsgd} transmits the gradient sign followed by the majority vote and \sketch \citep{rothchild2020fetchsgd,haddadpour2020fedsketch} uses a Count Mean Sketch to compress the gradients. We omit comparison with quantization methods \citep{vargaftik2022eden} given the difference in our objectives and the settings (\ref{eq:awgn_final}). \adsgd \citep{amiri2020machine} is a popular compression scheme for noisy channels, relying on Top-K and random sketching. However \adsgd does not scale to tasks of the size we consider and hence we benchmark against it only on \mnist. \ref{eq:sgd} serves as the noiseless baseline (\prettyref{tab:tasks}). All the compression algorithms use the error-feedback, and use the compression factor (compressed-gradient-size/original-size) $0.2$, the optimal in the range $[0.1, 0.8]$. We report the best results among $3$ independent runs for all the baselines (\S~\ref{app:add_exp}). 
% We defer additional details to \prettyref{app:add_exp}.

\subsection{Results on language modeling and image classification}
\label{sec:specific_results}

For \gpt language modeling, \prettyref{fig:llm} in \prettyref{sec:intro} highlights that \miss outperforms the baselines over a wide range of power levels. To the best of our knowledge, this is the first result of its kind to demonstrate gains for \gpt training over noisy channels. Specifically, we obtain $64\%$ improvement in perplexity over Z-SGD ($76$ vs. $212$) in the low power regime ($P = 10 \, \mathrm{K}$) and $50\%$ ($35$ vs. $71$) for the moderate one ($P=160\, \mathrm{K}$). This demonstrates the efficacy of \miss especially in the limited power environment. Indeed, \prettyref{tab:llm-power-ratio} illustrates that for a fixed target perplexity, \miss requires $16 \times$ less power than the second best, \ref{eq:awgn_sgd}. In the very high power regime, we observe no clear gains (as expected) compared to transmitting the uncompressed gradients directly via the \gsgd.

We observe a similar trend for \cifarten classification, as \prettyref{fig:cifar10} and \prettyref{tab:power-ratio} demonstrate the superiority of \miss over other compression schemes; \randk does better than the other baselines till moderate power levels after which \gsgd dominates. \signum is considerably worse than others, as it hasn't converged yet after $150$ epochs, and hence omitted. With regards to power reduction, \prettyref{tab:power-ratio} highlights that \miss requires just $(1/16)^{\text{th}}$ the power compared to \gsgd to reach any target accuracy till $91\%$. We observe similar gains for \cifarhun (\S~\ref{app:add_exp}).

% highlights that \miss outperforms the baselines on a wide range of power levels. In particular, we observe that \miss achieves accuracy gains of \nb{x} in the low-power regime and \nb{y} in the moderate region. In the high-power regime, all the baselines are close to the 	SGD baseline as the channel is almost noiseless. \nb{Figure 1b} further illustrates that for the same target accuracy, \miss requires just $(1/8)^{\text{th}}$ power compared to the baselines in the low-regime and a factor of $1/4$ in the moderate regime.

\prettyref{tab:mnist} compares the performance of \miss against various compression algorithms on \mnist. In the very noisy regime ($P=0.1$), \randk is slightly better than   \miss and outperforms the other baselines, whereas in the moderate ($P=1$) and high power ($P=10$) regimes, \miss is slightly better than the other algorithms. On the other hand, we observe that \adsgd performs worse than even simple compression schemes like \randk in all the settings.

\begin{table*}
  \centering
  \small
  \begin{minipage}[t]{0.52\textwidth}
    \caption{Test accuracy (\emph{higher the better}) after $50$ epochs on \mnist for low, moderate, and high power regimes.}   \centering
    \label{tab:mnist}%
    \vspace{2.7mm}
    \begin{tabularx}{1\linewidth}{Xlll}
    \toprule
    \multirow{2}{*}{\bf Algorithm} & \multicolumn{3}{c}{\bf Test accuracy} \\
    
    \cmidrule(lr){2-4}
   & $P=0.1$ & $P=1$ & $P=10$ \\
    \cmidrule(lr){1-1}\cmidrule(lr){2-4}
    Z-SGD & $81.3\%$ & $87.9\%$ & $91.9\%$ \\
    \cmidrule(lr){1-4}
    \signum & $76.7\%$ & $83.2\%$ & $85.4\%$ \\
    \randk & ${\bf 86.1\%}$  & $89.3\%$ & $91.5\%$ \\
    \sketch & $81.9\%$ & $88.2\%$ & $91.7\%$ \\
    \cmidrule(lr){1-4}
    \adsgd & $81.6\%$ & $86.9\%$ & $87.3\%$ \\

    \miss & $84.3\%$ & ${\bf 89.9 \% }$ & ${\bf 92.3\% }$ \\
    
    \bottomrule
\end{tabularx}
  \end{minipage}
\hspace{12pt}
  \begin{minipage}[t]{0.42\textwidth}
    % \scriptsize%
    \caption{Communication cost (\emph{lower the better}) for \gpt language modeling on \wikileaks. \miss transmits the lowest volume of data during training.}
    \centering 
    \label{tab:complexity}
    \vspace{1mm}
    \begin{tabularx}{1\linewidth}{Xll}
        \toprule
        {\bf Algorithm} &
        \multicolumn{2}{l}{\bf Data sent per iteration} \\
        
        \cmidrule(lr){1-3}

        Z-SGD & $496\,\mathrm{MB}$ & {\color{gray}($1 \times$)} \\
        \cmidrule(lr){1-3}
        
        \signum & $15\,\mathrm{MB}$  & {\color{gray}($33 \times$)} \\
        \randk & $99\,\mathrm{MB}$ &  {\color{gray}($5 \times$)} \\
        \sketch & $99\,\mathrm{MB}$ &  {\color{gray}($5 \times$)} \\
        \cmidrule(lr){1-3}
        \adsgd & $\mathrm{n/a}$ & $\mathrm{n/a}$ \\

        \miss & ${\bf 3\,\mathrm{\bf MB} }$ &  {\color{gray}($165 \times$)} \\
        
        \bottomrule
    \end{tabularx}
  \end{minipage}
  % \vspace{-3mm}
\end{table*}

\subsection{Power control: static vs. dynamic policies}
\label{sec:power-policy}

%\cite{saha2022}
\begin{figure}[t]
  \begin{center}
 \includegraphics[width=0.49\textwidth]{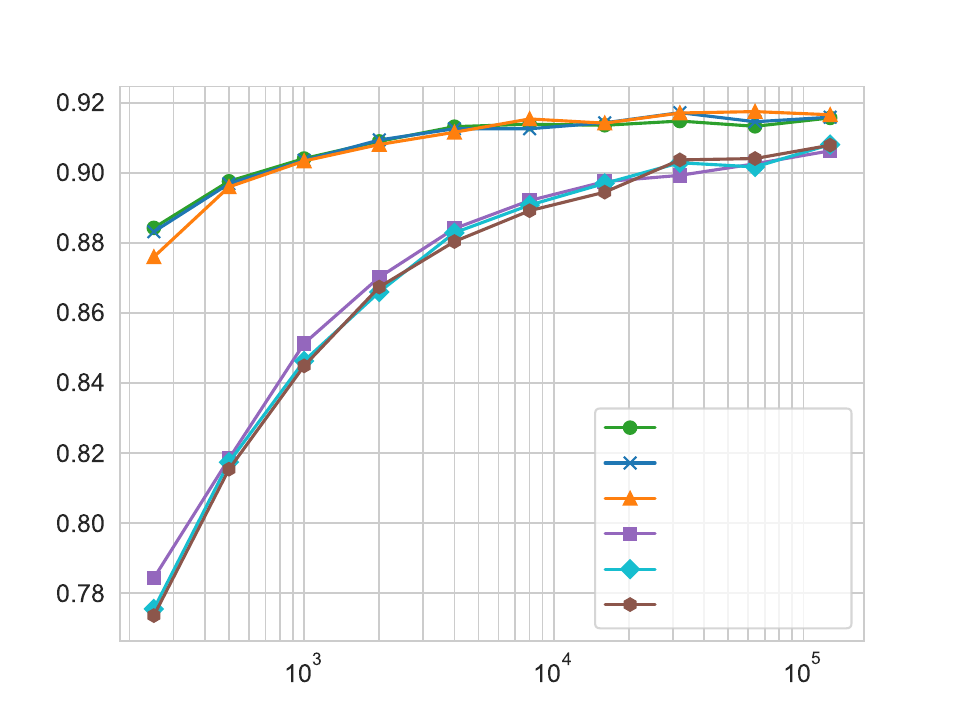}
    \put(-140,-2){\fontsize{7.5}{3}\selectfont Power budget $P$}
    \put(-238,80){\rotatebox[origin=t]{90}{\fontsize{7.5}{3}\selectfont Accuracy after 150 epochs}}
    \put(-72,71){\fontsize{6}{2}\selectfont LASER, constant}
    \put(-72,62){\fontsize{6}{2}\selectfont LASER, linear}
    \put(-72,53){\fontsize{6}{2}\selectfont LASER, step}
    \put(-72,44){\fontsize{6}{2}\selectfont Z-SGD, constant}
    \put(-72,35){\fontsize{6}{2}\selectfont Z-SGD, linear}
    \put(-72,26){\fontsize{6}{2}\selectfont Z-SGD, step}
 \end{center}
     \caption{Accuracy vs. budget $P$ for various laws. Constant is the best for both \miss and \gsgd. }
    \label{fig:power-policy}
    \vspace{-5mm}
\end{figure}
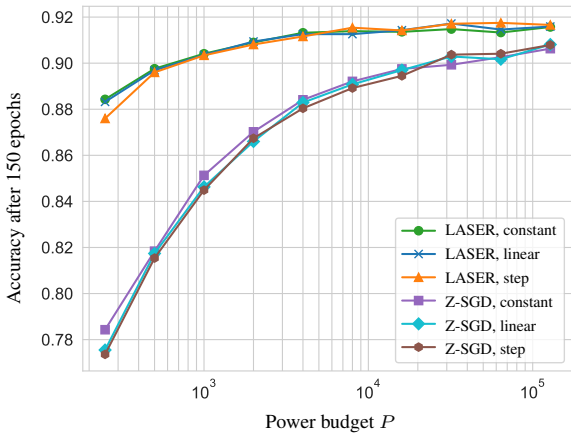
The formulation in \ref{eq:awgn_final} allows for any power control law $P_t$ as long as it satisfies the average power constraint: $\sum_t (P_t/T) \leq P$.
This begs a natural question: \emph{what's the best power scheme for LASER?}
To answer this, for \cifarten classification, under a fixed budget $P$ we consider different power policies  with both increasing and decreasing power across epochs: the constant, piecewise constant and linear schemes.
\prettyref{fig:power-policy} illustrates the results for the decreasing power laws, while \prettyref{fig:power-policy-increasing} their increasing counterparts. These results highlight that the \emph{constant} power policy achieves the \emph{best} performance for both LASER and Z-SGD, compared to the time-varying ones. Further LASER attains significant accuracy gains over Z-SGD for all the power control laws. Interestingly LASER performs the \emph{same} with all the power schemes. We posit this behavior to the fact that the \ref{eq:awgn_final} already contains a time-varying noise due to the term $\frac{\max_i \|\| \boldsymbol{g}_i \|\| }{\sqrt{P_t}}$. Since the gradients decay over time, this inherently allows for an implicit power/SNR-control law even with a constant $P_t$, thus enabling the constant power scheme to fare as good as the others. Hence, without loss of generality, we consider the static power schedule for our theory and experiments. We refer to \S~\ref{app:power-policy} for a detailed discussion.

\subsection{Computational complexity and communication cost}
\label{sec:complexity}
% Earlier results demonstrate the gains of \miss in terms of accuracy/perplexity and the power budget.
Recall from \prettyref{algo:miss_comm} that the two critical components of \miss are gradient compression and channel transmission. To gauge their efficacy we analyze them via two important metrics: (i) \emph{computational complexity} of compression and (ii) \emph{communication cost} of transmission. For (ii), recall from \prettyref{eq:normal_awgn} that the power constraint indirectly serves as a communication cost and encourages compression. \prettyref{tab:complexity} quantitatively measures the total data sent by clients for each training iteration (doesn't change with the power $P$) for \gpt language modeling on \wikileaks. As illustrated, \miss incurs the lowest communication cost among all the baselines with $165 \times$ cost reduction as compared to the Z-SGD, followed by \signum which obtains $33 \times$ reduction. Interestingly, \miss also achieves the best perplexity scores as highlighted in \prettyref{fig:llm}.  For these experiments, we let rank $r=4$ for \miss and the best compression factor $0.2$ for the baselines (as detailed earlier). \signum does not require any compression factor. For (i), since LASER relies on PowerSGD for the rank decomposition, it inherits the same low-complexity benefits: Tables $3$-$7$ of \citet{vogels2019powersgd} demonstrate that PowerSGD is efficient with significantly lower computational needs and has much smaller processing time/batch as compared to baselines without any accuracy drop. In fact, it is the core distributed algorithm behind the recent breakthrough DALL-E (\S~E in \citet{ramesh2021zero}).
\looseness=-1

% Note that the scalar gradient norms transmitted in \prettyref{algo:main} constitute less than $0.1$ MB of data. For example, for a target accuracy of $94.1 \% $ on \cifarten, \miss transmits , whereas ....

{\bf Slow and fast fading channels.} The slow/non-fading model in \prettyref{eq:normal_awgn} readily generalizes to the popular fast fading channel \citep{guo2020analog, amiri2020federated}: $\by = \sum_i \gamma_i \bx_i + \bZ$, where $\gamma_i$ are the channel fading coefficients. A standard technique here in the literature is to assume that channel-state-information (CSI) is known in the form of fading coefficients or their statistics, which essentially reduces the problem to a non-fading one. Likewise LASER can be extended to the fast fading channel as well. 
\looseness=-1

% The challenging setting without CSI is an interesting topic of future research.

% !TEX root = main.tex

% \section{Related work} 
% \label{sec:related} 

% We review the literature for lossy gradient compression along the themes (i) and (ii) from \prettyref{sec:intro}.  Further it performs well only for very high power settings and is worse than \randk in the regimes of interest (\prettyref{tab:mnist}).
{\bf Related work.} \textbf{(i) Compression schemes with noiseless communication.} Assuming a noiseless bit pipe from clients to the server, quantization methods \citep{dettmers20158,alistarh2017qsgd,horvoth2022natural,li2018network, wen2017terngrad,yu2019exploring, vargaftik2021drive} quantize each coordinate and send as fewer bits as possible. Sparsification techniques \citep{ivkin2019communication, stich2018sparsified, sun2019sparse,tsuzuku2018variance,wangni2018gradient} send a reduced number of coordinates, based on criteria such as Top/Random-K, as opposed to sending the full gradient directly. Hybrid methods \citep{dryden2016communication,lim20193lc} combine both. Rank compression methods \citep{yu2018gradiveq, cho2019gradzip, wang2018atomo} spectrally decompose gradient matrix (often via SVD) and transmit these factors. Since SVD is computationally prohibitive, we rely on the state-of-the-art light-weight compressor PowerSGD~\citep{vogels2019powersgd}. {\bf (ii) Compression schemes for noisy channels.}
The main idea here is to enable over-the-air-aggregation of gradients via the superposition nature of wireless channels~\citep{nazer2007computation} thus reducing the communication latency and bandwidth. The popular \adsgd~\citep{amiri2020machine} relies on Top-K sparsification and random sketching. However, being memory intensive, \adsgd is restricted to \mnist with $1$-layer NN and doesn't scale beyond. \citet{guo2020analog} propose an analog-gradient-aggregation scheme but it is limited to shallow neural networks. \citet{chang2020mac} design a digital quantizer for training over Gaussian MAC channels. {\bf (iii) Power laws.} In the absence of explicit power constraints, \citet{wei2022federated} show that $\mathcal{O}(1/t^2)$ noise-decay ensures the standard $1/T$ convergence rate for noisy FED-AVG whereas \citet{saha2022} propose a $t^{0.8}$ increase in SNR for the decentralized setup.  
\looseness=-1

\section{Conclusion}
\label{sec:concl}
We propose a principled gradient compression scheme, \miss, for wireless distributed optimization over additive noise channels. LASER attains significant gains over its baselines on a variety of metrics such as accuracy/perplexity, complexity and communication cost. It is an interesting avenue of future research to extend LASER to channels with downlink noise and fast fading without CSI.% , etc. 

\section*{Impact Statement}
This paper presents work whose goal is to advance the field of Machine Learning. There are many potential societal consequences of our work, none which we feel must be specifically highlighted here.

% By effectively leveraging the inherent low-rank structure of the gradients (and low-rank approximation combined with error feedback), \miss mitigates the noise in the communication and provably achieves a convergence rate comparable to the classical SGD for a noiseless setup. We empirically verified that LASER has much better evaluation metrics than the state of the art baselines, achieving $16 \times$ power saving on the distributed training of various neural architectures on language modeling and image classification benchmarks - over noisy channels. % ; LASER leads to the $10-100\times$ power saving compared to the baselines 
% % 
% % 
% % theory, empirical 
% % 
% Avenues for future research include considering other communication channels. While we considered the unweighted additive channels, we believe our techniques can be extended to the weighted ones, such as fading channels. Integrating current communication backends for distributed training like NCCL and GLOO with noisy channels is also an interesting topic of research.

% While we focused on the canonical AWGN channel in this paper, we believe our techniques can be generalized to other non-AWGN channels such as fading, T-distribution channels, etc., which is an interesting topic of future research.

%%%%%%%%%%%%%%%%%%%%%%%%%%%%%%%%%%%%%%%%%%%%%%%%
%% References
% \bibliographystyle{neurips_2022}
\bibliography{main}
\bibliographystyle{icml2024}

\clearpage

%%%%%%%%%%%%%%%%%%%%%%%%%%%%%%%%%%%%%%%%%%%%%%%%
%% Appendix
% !TEX root = main.tex
\newpage
\appendix
\onecolumn

{\bf Organization.} The appendix is organized as follows:

\begin{itemize}
    \item \prettyref{app:sgd_toolbox} contains the requisite material for error feedback and SGD convergence analysis. 
    \item \prettyref{app:miss_toolbox} details the important technical lemmas needed for the theoretical convergence of \miss.
    \item \prettyref{app:thm_proof} provides the proof for \prettyref{thm:main} whereas\prettyref{app:lmm_proof} contains the proofs of all technical lemmas.
    \item \prettyref{app:add_theory} provides additional details about the noisy channel and \prettyref{algo:miss_comm}.
    \item \prettyref{app:add_exp} contains additional experimental details and results.
\end{itemize}

\section{Error feedback and SGD convergence toolbox}
\label{app:sgd_toolbox}

In this section we briefly recall the main techniques for the convergence analysis of SGD with error feedback (\ref{eq:ef_sgd}) from \cite{stich2020error}. We consider $k=1$ clients with a compressor $\calC_r(\cdot)$ and without any channel communication noise $\awgn_P$(\prettyref{sec:background}):

\begin{align}
\begin{split}
    \btheta_{t+1} &= \btheta_t - \calC_r(\be_t + \gamma_t \bg_t)  \\
    \be_{t+1} &= (\be_t + \gamma_t \bg_t) - \calC_r(\be_t + \gamma_t \bg_t). 
\end{split}
    \tag{EF-SGD}
    \label{eq:ef_sgd}
\end{align}

Now we define the virtual iterates $\{\tilde{\btheta}_t \}_{t \geq 0}$ which are helpful for the convergence analysis: 
\begin{align}
    \tilde{\btheta}_t \define {\btheta}_t - \be_t.
    \label{eq:virtual_sgd}
\end{align}
Hence $\tilde{\btheta}_{t+1} = \btheta_t - \be_t - \gamma_t \bg_t = \tilde{\btheta}_t - \gamma_t \bg_t$. First we consider the case when $f$ is quasi-convex followed by the non-convex setting. In all the results below, we assume that the objective $f$ is $L$-smooth, gradient oracle $\bg$ has $(M, \sigma^2)$-bounded noise, and that $\calC_r(\cdot)$ satisfies the $\delta_r$ compression property (Assumptions~ \ref{assump:smooth}, \ref{assump:grad_oracle}, and \ref{assump:compressor}). 

\underline{\bf $f$ is quasi-convex:}

% for Lemmas \ref{lmm:quasi_progress}, \ref{lmm:quasi_errorbound}, and \ref{lmm:quasi_recursion}, and their corresponding result \prettyref{thm:quasi_result}. 
The following lemma gives a handle on the gap to optimality $\Expect{ \norm{\tilde \btheta_{t} - \btheta_\ast}^2 }$.

\begin{lemma}[{\cite{stich2020error}, Lemma 8}]
Let $\{\btheta_t, \be_t \}_{t \geq 0}$ be defined as in \ref{eq:ef_sgd}. Assume that $f$ is $\mu$-quasi convex for some $\mu \geq 0$. If $\gamma_t \leq \frac{1}{4L(1+M)}$ for all $t \geq 0$, then for $\{\tilde{\btheta}_t \}_{t \geq 0}$ defined in \prettyref{eq:virtual_sgd},
\begin{align}
    \Expect{ \norm{\tilde \btheta_{t+1} - \btheta_\ast}^2 } &\leq
 \left(1-\frac{\mu \gamma_t}{2}\right) \Expect{\norm{\tilde \btheta_{t} - \btheta_\ast}^2} 
 - \frac{\gamma_t}{2} \Expect{(f(\btheta_t)- f_\ast)} + \gamma_t^2 \sigma^2
 + 3 L \gamma_t   \Expect{\norm{\btheta_t - \tilde \btheta_t}^2}\,. 
 \label{eq:quasi_progress}
\end{align}
    \label{lmm:quasi_progress}
\end{lemma}

The following lemma bounds the squared norm of the error, \ie  $\Expect \norm{\be_{t}}^2 $, appearing in \prettyref{eq:quasi_progress}. Recall that a positive sequence $\iterates{a_t}$ is $\tau$-slow decreasing for parameter $\tau \geq 1$ if $a_{t+1} \leq a_t$ and $a_{t+1}(1+1/2\tau) \geq a_t$. The sequence $\iterates{a_t}$ is $\tau$-slow increasing if $\iterates{a_t^{-1}}$ is $\tau$-slow decreasing \cite{stich2020error}, Definition 10.

\begin{lemma}[{\cite{stich2020error}, Lemma 22}]
Let $\be_t$ be as in ~\eqref{eq:ef_sgd} for a $\delta_r$-approximate compressor $\rankcx$ and stepsizes $\{\gamma_t\}_{t \geq 0}$ with $\gamma_{t+1} \leq \frac{1}{10 L (2/\delta_r+M)}$, $\forall t \geq 0$ and $\{\gamma_t^2\}_{t \geq 0}$ $\frac{2}{\delta_r}$-slow decaying. Then
\begin{align}
 \Expect \qth{ 3L \norm{\be_{t+1}}^2 } \leq \frac{\delta_r}{64 L}\sum_{i=0}^t \left(1-\frac{\delta_r}{4}\right)^{t-i} \left(\Expect \norm{\nabla f(\btheta_{t-i})}^2 \right) +  \gamma_t  \sigma^2\,. \label{eq:sparse_bound1}
\end{align}
Furthermore, for any $\frac{4}{\delta_r}$-slow increasing non-negative sequence $\{w_t\}_{t \geq 0}$ it holds:
\begin{align*}
    3 L\sum_{t=0}^T w_t \Expect \norm{\be_t}^2 \leq \frac{1}{8L} \sum_{t=0}^T w_t \left(\Expect \norm{\nabla f(\btheta_{t})}^2 \right) +  \sigma^2 \sum_{t=0}^T w_t \gamma_t\,. %
\end{align*}
    \label{lmm:quasi_errorbound}
\end{lemma}

The following result controls the summations of the optimality gap that appear when combining \prettyref{lmm:quasi_progress} and \prettyref{lmm:quasi_errorbound}.

\begin{lemma}[{\cite{stich2020error}, Lemma 13}]
For every non-negative sequence $\{r_t\}_{t\geq 0}$ and any parameters $d \geq a > 0$, $c \geq 0$, $T \geq 0$, there exists a constant $\gamma \leq \frac{1}{d}$, such that for constant stepsizes $\{\gamma_t = \gamma\}_{t \geq 0}$ and weights $w_{t}:=(1-a\gamma)^{-(t+1)}$ it holds
\begin{align*}
\Psi_T := \frac{1}{W_T}\sum_{t=0}^T \left( \frac{w_t}{\gamma_t} \left(1-a \gamma_t \right) r_t - \frac{w_{t}}{\gamma_t} r_{t+1} + c \gamma_t w_t \right) = \tilde{\cal{O}} \left(d r_0 \exp\left[- \frac{aT}{d} \right] + \frac{c}{aT}  \right)\,.
\end{align*}
    \label{lmm:quasi_recursion}
\end{lemma}

Combining the above lemmas, we obtain the following result for the convergence rate of \ref{eq:ef_sgd}.

\begin{theorem}[{\cite{stich2020error}, Theorem 22}]
\label{thm:quasi_result}
Let $\{\btheta_t\}_{t \geq 0}$ denote the iterates of the error compensated stochastic gradient descent~\eqref{eq:ef_sgd} 
with constant stepsize $\{\gamma_t = \gamma\}_{t \geq 0}$ and
with a $\delta_r$-approximate compressor on a differentiable function $f \colon \reals^d \to \reals$ under Assumptions~\ref{assump:smooth} and~\ref{assump:grad_oracle}. Then, if $f$ 
\begin{itemize}[nosep,leftmargin=12pt,itemsep=2pt]
 \item satisfies Assumption~\ref{assump:quasi} for $\mu > 0$, then there exists a stepsize $\gamma \leq \frac{1}{10L(2/\delta_r + M)}$ (chosen as in Lemma~\ref{lmm:quasi_recursion})
such that
 \begin{align*}
 \end{align*}
 where the output $\btheta_{\rm out} \in \{\btheta_t\}_{t=0}^{T-1}$ is chosen to be $\btheta_t$ with probability proportional to 
 $(1-\mu \gamma/2)^{-t}$. 
 \item satisfies Assumption~\ref{assump:quasi} for $\mu = 0$, then there exists a stepsize $\gamma \leq \frac{1}{10L(2/\delta_r + M)}$ (chosen as in Lemma~\ref{lmm:quasi_recursion}) such that
 \begin{align*}
 \Expect f(\btheta_{\rm out})-f_* = \calO \left(  \frac{L(1/\delta_r + M) \norm{\btheta_0-\btheta_\ast}^2 }{ T } + \frac{\sigma \norm{\btheta_0-\btheta_\ast} }{\sqrt T}  \right)\,,
 \end{align*}
  where the output $\btheta_{\rm out} \in \{\btheta_t\}_{t=0}^{T-1}$ is chosen uniformly at random from the iterates $\{\btheta_t\}_{t=0}^{T-1}$.
\end{itemize}

\end{theorem}

\underline{\bf $f$ is non-convex:}

Now we consider the case where $f$ is an arbitrary non-convex function. The above set of results extend in a similar fashion to this setting too as described below:

\begin{lemma}[{\cite{stich2020error}, Lemma 9}]
Let $\{\btheta_t, \be_t \}_{t \geq 0}$ be defined as in \ref{eq:ef_sgd}. If $\gamma_t \leq \frac{1}{2L(1+M)}$ for all $t \geq 0$, then for $\{\tilde{\btheta}_t \}_{t \geq 0}$ defined in \prettyref{eq:virtual_sgd},
\begin{align}
  \Expect[f(\tilde \btheta_{t+1})] \leq \Expect [f(\tilde \btheta_t)] - \frac{\gamma_t}{4}\Expect \norm{\nabla f(\btheta_t)}^2 +\frac{\gamma_t^2 L \sigma^2}{2} + \frac{\gamma_t L^2}{2}\Expect \norm{\btheta_t - \tilde \btheta_t}^2\,.
 \label{eq:noncvx_progress}
\end{align}
    \label{lmm:noncvx_progress}
\end{lemma}

\begin{lemma}[{\cite{stich2020error}, Lemma 22}]
Let $\be_t$ be as in ~\eqref{eq:ef_sgd} for a $\delta_r$-approximate compressor $\rankcx$ and stepsizes $\{\gamma_t\}_{t \geq 0}$ with $\gamma_{t+1} \leq \frac{1}{10 L (2/\delta_r+M)}$, $\forall t \geq 0$ and $\{\gamma_t^2\}_{t \geq 0}$ $\frac{2}{\delta_r}$-slow decaying. Then
\begin{align}
 \Expect \qth{ 3L \norm{\be_{t+1}}^2 } \leq \frac{\delta_r}{64 L}\sum_{i=0}^t \left(1-\frac{\delta_r}{4}\right)^{t-i} \left(\Expect \norm{\nabla f(\btheta_{t-i})}^2 \right) +  \gamma_t  \sigma^2\,. \label{eq:sparse_bound2}
\end{align}
Furthermore, for any $\frac{4}{\delta_r}$-slow increasing non-negative sequence $\{w_t\}_{t \geq 0}$ it holds:
\begin{align*}
    3 L\sum_{t=0}^T w_t \Expect \norm{\be_t}^2 \leq \frac{1}{8L} \sum_{t=0}^T w_t \left(\Expect \norm{\nabla f(\btheta_{t-i})}^2 \right) +  \sigma^2 \sum_{t=0}^T w_t \gamma_t\,. %
\end{align*}
    \label{lmm:noncvx_errorbound}
\end{lemma}

\begin{lemma}[{\cite{stich2020error}, Lemma 14}]
For every non-negative sequence $\{r_t\}_{t\geq 0}$ and any parameters $d \geq 0$, $c \geq 0$, $T \geq 0$, there exists a constant $\gamma \leq \frac{1}{d}$, such that for constant stepsizes $\{\gamma_t = \gamma\}_{t \geq 0}$ it holds:
\begin{align*}
\Psi_T  := \frac{1}{T+1} \sum_{t=0}^T \left( \frac{r_t}{\gamma_t} - \frac{r_{t+1}}{\gamma_t} + c \gamma_t \right) \leq \frac{d r_0}{T+1} + \frac{2\sqrt{c r_0}}{\sqrt{T+1}} \,.
\end{align*}
    \label{lmm:noncvx_recursion}
\end{lemma}

Now we have the final convergence result for the non-convex setting.
\begin{theorem}[{\cite{stich2020error}, Theorem 22}]
Let $\{\btheta_t\}_{t \geq 0}$ denote the iterates of the error compensated stochastic gradient descent~\eqref{eq:ef_sgd} 
with constant stepsize $\{\gamma_t = \gamma\}_{t \geq 0}$ and
with a $\delta_r$-approximate compressor on a differentiable function $f \colon \reals^d \to \reals$ under Assumptions~\ref{assump:smooth} and~\ref{assump:grad_oracle}. Then, if $f$ is an arbitrary non-convex function, there exists a stepsize $\gamma \leq \frac{1}{10L(1/\delta_r + M)}$ (chosen as in Lemma~\ref{lmm:noncvx_recursion}), such that
  \begin{align*}
 \Expect \norm{\nabla f(\btheta_{\rm out})}^2 = \calO \left(  \frac{L(1/\delta_r + M) (f(\btheta_0) - f_\ast) }{ T } + \sigma \sqrt{\frac{L(f(\btheta_0) - f_\ast)}{T}}  \right)\,.
 \end{align*}
 where the output $\btheta_{\rm out} \in \{\btheta_t\}_{t=0}^{T-1}$ is chosen uniformly at random from the iterates $\{\btheta_t\}_{t=0}^{T-1}$.
\label{thm:noncvx_result}
\end{theorem}

\section{Technical lemmas for \miss convergence}
\label{app:miss_toolbox}

Towards the convergence analysis of \miss for $k=1$, we rewrite the \prettyref{algo:miss_comm} succinctly as:
\begin{align}
\begin{split}
    \btheta_{t+1} &= \btheta_t - \awgn_{(\balpha, \bbeta)}\pth{\calC_r(\be_t + \gamma_t \bg_t)}  \\
    \be_{t+1} &= (\be_t + \gamma_t \bg_t) - \calC_r(\be_t + \gamma_t \bg_t)\, ,
\end{split}
    \tag{\textsc{LASER}}
    \label{eq:laser_rewri}
\end{align}
where the channel corrupted gradient approximation $\awgn_{(\balpha, \bbeta)}(\cdot)$ is given by
\begin{align}
    \awgn_{(\balpha, \bbeta)}(\underbrace{\calC_r(\be_t + \gamma_t \bg_t)}_{=\bP \bQ^\top}) \define \sum_{i=1}^r \pth{\bp_i + \frac{\norm{\bp_i}}{\sqrt{\alpha_i}} \cdot \bZ_m^{(i)}} \pth{\bq_i + \frac{\norm{\bq_i}}{\sqrt{\beta_i}} \cdot \bZ_n^{(i)}}^\top,
    \label{eq:miss_reconstr}
\end{align}
and $\balpha=(\alpha_i)_{i=1}^r$ and $\bbeta=(\beta_i)_{i=1}^r$ are appropriate power allocations to transmit the respective left and right factors $\bP=[\bp_1, \ldots, \bp_r] \in \matrx{m}{r}$ and $\bQ=[\bq_1, \ldots, \bq_r] \in \matrx{n}{r}$ for the decomposition $\calC_r(\be_t + \gamma_t \bg_t) = \bP \bQ^\top$. $\bZ_{m}^{(i)} \in \reals^m$ and $\bZ_{n}^{(i)} \in \reals^n$ denote the independent channel noises for each factor $i \in [r]$.

% of the decomposition $\calC_r(\be_t + \gamma_t \bg_t) = \bP \bQ^\top, \, \bP \in \matrx{m}{r}, \bQ \in \matrx{n}{r}$, and 

Thus we observe from \ref{eq:laser_rewri} that it has an additional channel corruption in the form of $\awgn_{(\balpha, \bbeta)}(\cdot)$ as compared to the \ref{eq:ef_sgd}. Now in the remainder of this section, we explain how to choose the power allocation $(\balpha, \bbeta)$ (\prettyref{app:power_alloc}), how to control the influence of the channel $\awgn_{(\balpha, \bbeta)}(\cdot)$ on the convergence of \miss (\prettyref{app:channel_influ}), and utilize these results to establish technical lemmas along the lines of \prettyref{app:sgd_toolbox} for \miss (\prettyref{app:onestep_progr}).

\subsection{Power allocation}
\label{app:power_alloc}
In this section, we introduce the key technical lemmas about power allocation that are crucial for the theoretical results. We start with the rank one case.

\begin{lemma}[\bf Rank-$1$ power allocation]
\label{lmm:power_rankone}
For a power $P >0$ and $m,n \in \naturals$ with $m \leq n$, define the function $f_P:\reals_{+} \times \reals_{+} \to \reals_{+}$ as
\begin{align*}
    f_P(\alpha, \beta ) \define \pth{1+ \frac{m} {\alpha}} \pth{1+ \frac{n}{\beta}},
\end{align*}
and the constraint set $S_{P} \define  \sth{(\alpha, \beta ): \alpha \geq 0, \beta \geq 0, \alpha + \beta = P }$. Then for the minimizer $(\alpha^\ast, \beta^\ast) = \argmin_{(\alpha, \beta) \in S_P} f_P(\alpha, \beta)$, we have
\begin{align*}
    f_P(\alpha^\ast, \beta^\ast) \leq 1 + \frac{4}{m \,\snr}\pth{1+ \frac{1}{n \, \snr}}, \quad \snr \define \frac{P}{mn}\, .
\end{align*}
Further the minimizer is given by
\begin{align*}
    \alpha^\ast &= \begin{cases} \sqrt{1+\frac{P}{n}} \pth{\frac{\sqrt{1+\frac{P}{m}}-\sqrt{1+\frac{P}{n}}}{\frac{1}{m}-\frac{1}{n}}}, & m \neq n \\
    P/2, & m=n
    \end{cases} \\
    \beta^\ast & = P- \alpha^\ast.
\end{align*}
\end{lemma}

\begin{lemma}[\bf Rank-$r$ power allocation]
\label{lmm:power_rankr}
For a power $P >0$,  $m,n, r \in \naturals$ with $m \leq n$, and positive scalars $\kappa_1, \ldots, \kappa_r > 0$ with $\sum_i \kappa_i = 1$, define the function $f_P:(\reals_{+})^r \times (\reals_{+} )^r \to \reals_{+}$ as
\begin{align*}
    f_P (\balpha, \bbeta) \define \sum_{i=1}^r \kappa_i \pth{1+ \frac{m} {\alpha_i}} \pth{1+ \frac{n}{\beta_i}}, \quad \balpha = (\alpha_i)_{i=1}^r, \, \bbeta = (\beta_i)_{i=1}^r, 
\end{align*}
and the constraint set $S_{P} \define  \sth{ \{(\balpha, \bbeta) : \balpha \geq 0, \bbeta \geq 0, \sum_i (\alpha_i + \beta_i) = P }$. Then there exists a power allocation scheme $(\balpha^\ast, \bbeta^\ast) \in S_P$ such that
\begin{align*}
    \min_{(\balpha, \bbeta) \in S_P} f_P(\balpha, \bbeta) \leq f_P(\balpha^\ast, \bbeta^\ast) \leq 1 + \frac{4}{(m/r) \, \snr}\pth{1+ \frac{1}{(n/r) \,  \snr}},
\end{align*}
where $\snr \define \frac{P}{mn}$. Further $(\balpha^\ast, \bbeta^\ast)$ is given by 
\begin{align*}
    \alpha_i^\ast &= \begin{cases} \sqrt{1+\frac{P_i}{n}} \pth{\frac{\sqrt{1+\frac{P_i}{m}}-\sqrt{1+\frac{P_i}{n}}}{\frac{1}{m}-\frac{1}{n}}}, & m \neq n \\
    P_i/2, & m=n
    \end{cases} \\
    \beta_i^\ast & = P_i - \alpha_i^\ast,\\
    P_i & = P \pth{\frac{\sqrt{\kappa_i}}{\sum_j \sqrt{\kappa_j}}}. 
\end{align*}
\end{lemma}
\begin{remark}
\normalfont
In other words, we first divide the power $P$ proportional to $\sqrt{\kappa_i}$ for each $i \in [r]$ and further allocate this $P_i$ amongst $\alpha_i^\ast$ and $\beta_i^\ast$ as per the optimal rank one allocation scheme in \prettyref{lmm:power_rankone}. 
\end{remark}

\subsection{Channel influence factor}
\label{app:channel_influ}
In this section we establish the bounds for the channel influence defined in \prettyref{eq:chanel_infl} for both \gsgd and \miss. This helps us give a handle to control the second moment of the gradient corrupted by channel noise.

\begin{lemma}[\bf Channel influence on \gsgd] 
For the \ref{eq:awgn_sgd} algorithm that sends the uncompressed gradients directly over the \ref{eq:awgn_final} with power constraint $P$, we have
\begin{align}
    \lambda_\gsgd = \frac{1}{\snr},
\end{align}
where $\snr = \frac{P}{mn} $.
    \label{lmm:sgd_influ}
\end{lemma}

\begin{lemma}
For the \ref{eq:laser_rewri} algorithm with the optimal power allocation $(\balpha, \bbeta)$ (chosen as in \prettyref{lmm:power_rankr}), we have
\begin{align}
    \lambda_\miss \leq  \frac{4}{(m/r) \, \snr}\pth{1+ \frac{1}{(n/r) \, \snr}},
\end{align}
where $\snr = \frac{P}{mn} $.
    \label{lmm:miss_influ}
\end{lemma}
\begin{remark}
\normalfont
Note that for the optimal power allocation via \prettyref{lmm:power_rankr}, we need the positive scalars $\kappa_1, \ldots, \kappa_r$. In the context of \miss, we will later see in the proof in \prettyref{app:lmm_proof} that $\kappa_i \propto \norm{\bp_i}^2 $.
\end{remark}

Thus \prettyref{lmm:sgd_influ} and \prettyref{lmm:miss_influ} establish that
\begin{align*}
    \lambda_\miss \leq  \frac{4}{(m/r) \, \snr} \pth{1+ \frac{1}{(n/r) \, \snr}} \ll \frac{1}{\snr} = \lambda_{\gsgd}.
\end{align*}
In the low-rank \cite{vogels2019powersgd} and constant-order SNR regime where $r = \bigo(1)$ and $\snr = \Omega(1)$, we observe that $\lambda_\miss$ is roughly $\bigo(m)$ times smaller than $\lambda_{\gsgd}$.

{\bf Note on assumption between $\missinflu$ and $\delta_r$.} Recall from \ref{eq:laser_rewri} that the local memory $\be_t$ has only access to the compressed gradients and not the channel output. In an hypothetical scenario, where it has access to the same, it follows that $\Expect_{\bZ} \norm{\awgn_{(\balpha, \bbeta)}(\rankcx(\bM)) - \bM}^2 \leq (1- (\delta_r - \missinflu)) \norm{\bM}^2$. Hence for the compression property in this ideal scenario, we need $\lambda_\miss \leq \delta_r$.

\subsection{Optimality gap and error bounds for ~\ref{eq:laser_rewri} iterates}
\label{app:onestep_progr}
In this section, we characterize the gap to the optimality and the error norm for the \ref{eq:laser_rewri} iterates $\iterates{\btheta_t}$ (similar to Lemmas ~\ref{lmm:quasi_progress}, \ref{lmm:quasi_errorbound}, \ref{lmm:quasi_errorbound} and \ref{lmm:noncvx_errorbound} for \ref{eq:ef_sgd}). Towards the same, first we define the virtual iterates $\iterates{\tilde{\btheta_t}}$ as follows:
\begin{align}
 \tilde{\btheta}_t \define {\btheta}_t - \be_t \,.  
 \label{eq:virtual_miss}
\end{align}
Thus,
\begin{align}
 \tilde{\btheta}_{t+1} = {\btheta}_{t+1} - \be_{t+1} = \tilde{\btheta}_t -\gamma_t \bg_t + \calC_r(\be_t + \gamma_t \bg_t) - \awgn_{(\balpha, \bbeta)}\pth{\calC_r(\be_t + \gamma_t \bg_t)}.
 \label{eq:virtualrelative_miss}
\end{align}

The following lemma controls the optimality gap $\Expect \norm{\tilde{\btheta}_t - \btheta_\ast}^2$ when $f$ is quasi-convex.
\begin{lemma}[\bf Descent for quasi-convex]
Let $\{\btheta_t, \be_t \}_{t \geq 0}$ be defined as in \ref{eq:laser_rewri}. Assume that $f$ is $\mu$-quasi convex for some $\mu \geq 0$ and that Assumptions~\ref{assump:smooth} and \ref{assump:grad_oracle} hold. If $\gamma_t \leq \frac{1}{4L(1+M)}\, \pth{\frac{1-2\lambda_\miss}{1+\lambda_\miss}}$ for all $t \geq 0$, then for $\{\tilde{\btheta}_t \}_{t \geq 0}$ defined in \prettyref{eq:virtual_miss},
\begin{align}
\begin{split}
    \Expect{ \norm{\tilde \btheta_{t+1} - \btheta_\ast}^2 } &\leq
 \left(1-\frac{\mu \gamma_t}{2}\right) \Expect{\norm{\tilde \btheta_{t} - \btheta_\ast}^2} 
 - \frac{\gamma_t}{2} \Expect{(f(\btheta_t)- f_\ast)} + \gamma_t^2 \sigma^2(1+\lambda_\miss) \\
 & \hspace{9em}
 + (3 L \gamma_t (1+\lambda_\miss) + \lambda_\miss   ) \Expect{\norm{\btheta_t - \tilde \btheta_t}^2}\,. 
\end{split} 
 \label{eq:missquasi_progress}
\end{align}
    \label{lmm:onestep_miss}
\end{lemma}

Notice that \prettyref{lmm:onestep_miss} is similar to \prettyref{lmm:quasi_progress} for noiseless EF-SGD except for an additional channel influence factor $\lambda_\miss$. The following result bounds the error norm.

\begin{lemma}[\bf Error control]
Let $\be_t$ be as in ~\eqref{eq:laser_rewri} for a $\delta_r$-approximate compressor $\rankcx$ and stepsizes $\{\gamma_t\}_{t \geq 0}$ with $\gamma_{t} \leq \frac{1}{10 L (2/\delta_r+M)(1+\lambda_\miss)}$, $\forall t \geq 0$ and $\{\gamma_t^2\}_{t \geq 0}$ $\frac{2}{\delta_r}$-slow decaying. Further suppose that Assumption~\ref{assump:influence} holds. Then
\begin{align}
\begin{split}
 \pth{3L(1+\lambda_\miss)+\frac{\lambda_\miss}{\gamma_t}}\Expect \norm{\be_{t+1}}^2 &\leq \frac{\delta_r}{32 L}\sum_{i=0}^t \left(1-\frac{\delta_r}{4}\right)^{t-i} \left(\Expect \norm{\nabla f(\btheta_{t-i})}^2 \right) \\
 & \hspace{8em}
 +  \gamma_t  \sigma^2(1+\lambda_\miss) \,. 
\end{split}
\label{eq:misserror_control}
\end{align}
Furthermore, for any $\frac{4}{\delta_r}$-slow increasing non-negative sequence $\{w_t\}_{t \geq 0}$ it holds:
\begin{align}
\begin{split}
    \pth{3L(1+\lambda_\miss)+\frac{\lambda_\miss}{\gamma_t}} \sum_{t=0}^T w_t \Expect \norm{\be_t}^2 &\leq \frac{1}{6L} \sum_{t=0}^T w_t \left(\Expect \norm{\nabla f(\btheta_{t})}^2 \right)  \\
    & \hspace{1em} 
    + \sigma^2(1+\lambda_\miss) \sum_{t=0}^T w_t \gamma_t\,. %
\end{split}
\label{eq:misserrorsum_control}
\end{align}
    \label{lmm:misserror_control}
\end{lemma}

The following lemma establishes the progress in the descent for non-convex case.
\begin{lemma}[\bf Descent for non-convex]
Let $\{\btheta_t, \be_t \}_{t \geq 0}$ be defined as in \ref{eq:laser_rewri} and that Assumptions~\ref{assump:smooth} and \ref{assump:grad_oracle} hold. If $\gamma_t \leq \frac{1}{4L(1+M)(1+\lambda_\miss)}$ for all $t \geq 0$, then for $\{\tilde{\btheta}_t \}_{t \geq 0}$ defined in \prettyref{eq:virtual_miss},
\begin{align}
\begin{split}
  \Expect[f(\tilde \btheta_{t+1})] &\leq \Expect [f(\tilde \btheta_t)] - \frac{\gamma_t}{4}\Expect \norm{\nabla f(\btheta_t)}^2 +\frac{\gamma_t^2 L \sigma^2(1+\lambda_\miss)}{2} \\
  &\hspace{7em} + \Expect \norm{\btheta_t - \tilde \btheta_t}^2 \pth{\frac{L^2 \gamma_t}{2} + L \lambda_\miss}.
 \label{eq:missnoncvx_progress}
\end{split}
\end{align}
    \label{lmm:missnoncvx_progress}
\end{lemma}

\section{Proof of \prettyref{thm:main}}
\label{app:thm_proof}

\begin{proof}
We prove the bounds in (i) and (ii) when $f$ is quasi-convex, (iii) when $f$ is an arbitrary non-convex function, and (iv) for \gsgd.

\underline{\bf (i), (ii) $f$ is $\mu$-quasi-convex:}
Observe that the assumptions of \prettyref{thm:main} automatically satisfy the conditions of \prettyref{lmm:onestep_miss}. Denoting $r_t \define \Expect{ \norm{\tilde \btheta_{t+1} - \btheta_\ast}^2 }$ and $s_t \define \Expect{(f(\btheta_t)- f_\ast)}$, for any $w_t >0$ we obtain
\begin{align*}
    \frac{w_t}{2} s_t \stackrel{\eqref{eq:missquasi_progress}}{\leq} \frac{w_t}{\gamma_t}\pth{1-\frac{\mu \gamma_t}{2}} r_t  -\frac{w_t}{\gamma_t} r_{t+1} + \gamma_t w_t \sigma^2(1+\missinflu) + 3 w_t (L(1+\missinflu)+\frac{\missinflu}{\gamma_t}) \Expect \norm{\be_t}^2 \, .
\end{align*}
Taking summation on both sides and invoking Lemma~\ref{lmm:quasi_errorbound} (assumption on $w_t$ verified below), 
\begin{align*}
    \sum_{t=0}^T \frac{w_t}{2} s_t \stackrel{\eqref{eq:misserrorsum_control}}{\leq} \sum_{t=0}^T \pth{ \frac{w_t}{\gamma_t}\pth{1-\frac{\mu \gamma_t}{2}} r_t  -\frac{w_t}{\gamma_t} r_{t+1} + 2 \gamma_t w_t \sigma^2(1+\missinflu)} +
\frac{1}{6L} \sum_{t=0}^T w_t \left(\Expect \norm{\nabla f(\btheta_{t})}^2 \right). 
\end{align*}
Since $f$ is $L$-smooth, we have $\norm{\nabla f(\btheta_t)}^2 \leq 2L(f(\btheta_t) - f_\ast)$. Now rewriting the above inequality, we have
\begin{align*}
    \frac{1}{6} \sum_{t=0}^T  w_t s_t \leq \sum_{t=0}^T \pth{ \frac{w_t}{\gamma_t}\pth{1-\frac{\mu \gamma_t}{2}} r_t  -\frac{w_t}{\gamma_t} r_{t+1} + 2 \gamma_t w_t \sigma^2(1+\missinflu)}. 
\end{align*}
Substituting $W_T \define \sum_{t=0}^T w_t$,
\begin{align*}
    \frac{1}{W_T} \sum_{t=0}^T  w_t s_t \leq  \frac{6}{W_T} \sum_{t=0}^T \pth{ \frac{w_t}{\gamma_t}\pth{1-\frac{\mu \gamma_t}{2}} r_t  -\frac{w_t}{\gamma_t} r_{t+1} + 2 \gamma_t w_t \sigma^2(1+\missinflu)} =: \Xi_T \,.
\end{align*}

Now it remains to derive the estimate for $\Xi_T$. Towards this, (i) if $\mu>0$ and with constant stepsize $\gamma_t = \gamma \leq \frac{1}{10L(\frac{2}{\delta_r}+M)(1+\missinflu)}$, we observe that $(1-\frac{\mu \gamma}{2}) \geq \pth{1- \frac{\delta_r}{16}} $ and by Example 1 in \cite{stich2020error}, the weights $w_t = \pth{1 - \frac{\mu \gamma}{2}}^{-(t+1)}$ are $2 \tau$-slow increasing with $\tau= \frac{2}{\delta_r}$. Hence the claim in (i) follows by applying Lemma~\ref{lmm:quasi_recursion} and observing that the sampling probablity to choose $\btheta_{\rm out}$ from $\{ \btheta_t \}_{t=0}^{T-1}$ is same as $w_t$. 

For (ii) with constant stepsize and $\mu=0$, we apply \prettyref{lmm:noncvx_recursion} by setting the weights $w_t=1$. 

\underline{\bf (iii) $f$ is non-convex}
The proof in this case is very similar to that of the above. Denoting $r_t \define 4 \Expect[f(\tilde \btheta_{t}) - f_\ast], s_t \define
\Expect \norm{\nabla f(\btheta_t)}^2, c = 4L\sigma^2(1+\missinflu)$, and $w_t=1$, we have from \prettyref{lmm:missnoncvx_progress} that
\begin{align*}
    \frac{s_t}{4} \stackrel{\eqref{eq:missnoncvx_progress}}{\leq} \frac{r_t}{4 \gamma_t}  - \frac{r_{t+1}}{4 \gamma_t} +  \frac{\gamma_t c}{8} + L \pth{\frac{L}{2} + \frac{\lambda_\miss}{\gamma_t}} \Expect \norm{\be_t}^2.
\end{align*}
Since $\frac{L}{2} \leq 3L(1+\missinflu)$, multiplying both sides of the above inequality by $w_t$ and taking summation, we obtain
\begin{align*}
    \frac{1}{4W_T} \sum_{t=0}^T w_t s_t \stackrel{\eqref{eq:misserrorsum_control}}{\leq} \frac{1}{W_T} \sum_{t=0}^T w_t \pth{ \frac{r_t}{4 \gamma_t} - \frac{r_{t+1}}{4 \gamma_t} + \frac{\gamma_t c}{8} } + \frac{L}{W_T} \pth{\sum_{t=0}^T \frac{w_t s_t}{6L}+ \frac{c w_t \gamma_t}{4 L} },
\end{align*}
which upon rearranging gives
\begin{align*}
    \frac{1}{W_T} \sum_{t=0}^T w_t s_t {\leq} \frac{12}{W_T} \sum_{t=0}^T w_t \pth{ \frac{r_t}{4 \gamma_t} - \frac{r_{t+1}}{4 \gamma_t} + \frac{3\gamma_t c}{8} }.
\end{align*}
Now invoking \prettyref{lmm:noncvx_recursion} yields the final result in (iii).

\underline{\bf \gsgd:} Recall from ~\ref{eq:awgn_sgd} that the iterates $\iterates{\btheta_t}$ are given by
\begin{align*}
    \btheta_{t+1} = \btheta_t - \gamma_t \, \awgnp(\bg_t ).
\end{align*}
Thus \gsgd can be thought of as a special case of \ref{eq:ef_sgd} with no compression, \ie $\delta_r =1$, and hence we can utilize the same convergence tools. It remains to estimate the first and second moments of the stochastic gradient $\awgnp(\bg_t )$. Recall from the definition of $\awgnp$ in the ~\ref{eq:awgn_final} that $\awgnp(\bg_t ) = \bg_t + \frac{\norm{\bg_t}}{\sqrt{P}}\, \bZ_t$, where $\bZ_t$ is a zero-mean independent channel noise, and from \prettyref{assump:grad_oracle} that $\bg_t = \nabla f(\btheta_t) + \bxi_t $ with a $(M, \sigma^2)$-bounded noise $\bxi_t$. Hence
\begin{align*}
   \Expect\qth{\awgnp(\bg_t )|\btheta_t} &=  \Expect[\bg_t |\btheta_t] = \nabla f(\btheta_t), \\
   \Expect\qth{\norm{\awgnp(\bg_t ) - \nabla f(\btheta_t)}^2|\btheta_t} &= \Expect\qth{\norm{\awgnp(\bg_t ) - \bg_t + \bg_t - \nabla f(\btheta_t)}^2|\btheta_t} \\
   & = \Expect \qth{\norm{\awgnp(\bg_t ) - \bg_t}^2|\btheta_t} + \Expect \qth{\norm{\bg_t - \nabla f(\btheta_t)}^2|\btheta_t} \\
  & \stackrel{~\ref{eq:chanel_infl}}{=} \Expect \qth{\lambda_\gsgd \norm{\bg_t}^2 | \btheta_t} + \Expect \norm{\bxi_t}^2 \\
  & = \lambda_\gsgd \norm{\nabla f(\btheta_t)}^2 + (1+ \lambda_\gsgd)\Expect \norm{\bxi_t}^2 \\
  & \leq (M+1)(1+ \lambda_\gsgd) \norm{\nabla f(\btheta_t)}^2 + (1+ \lambda_\gsgd) \sigma^2.
\end{align*}

Thus \gsgd satifies the $(\tilde M, \tilde{\sigma}^2)$-bounded noise condition in \prettyref{assump:grad_oracle} with $\tilde M = (M+1)(1+ \lambda_\gsgd) $ and $\tilde{\sigma}^2 = (1+ \lambda_\gsgd) \sigma^2$. Thus the claim (iv) follows from applying \prettyref{thm:quasi_result} and \prettyref{thm:noncvx_result} with the constants $\delta_r \rightarrow 1, M \rightarrow \tilde M, \sigma^2 \rightarrow \tilde{\sigma}^2$.

Finally, \prettyref{lmm:sgd_influ} and \prettyref{lmm:miss_influ} establish the relation between the channel influence factors $\lambda_\gsgd$ and $\missinflu$.

\end{proof}

\section{Proof of technical lemmas}
\label{app:lmm_proof}

\subsection{Proof of \prettyref{lmm:power_rankone}}

\begin{proof}
    Since $\log(\cdot)$ is a monotonic function, minimizing $f_P(\alpha, \beta)$ over $S_{P} = \sth{(\alpha, \beta ): \alpha \geq 0, \beta \geq 0, \alpha + \beta = P }$ is equivalent to minimizing $\log f_P(\alpha, \beta) = \log \pth{1 + \frac{m}{\alpha}} + \log \pth{1 + \frac{n}{\beta}}$. Define the Lagrangian $L(\alpha, \beta, \lambda)$ as
    \begin{align*}
        L(\alpha, \beta, \lambda) \define  \log \pth{1 + \frac{m}{\alpha}} + \log \pth{1 + \frac{n}{\beta}} + \lambda(\alpha+ \beta - P).
    \end{align*}
    Letting $\nabla_\alpha L = \nabla_\beta L =0$, we obtain that $\frac{m}{\alpha(m+\alpha)} = \frac{n}{\beta(n+\beta)}$. Now constraining $\alpha+ \beta = P$, we obtain the following quadratic equation:
    \begin{align*}
        \alpha^2 \pth{\frac{1}{m}- \frac{1}{n}} + 2 \alpha \pth{1+ \frac{P}{n}} - \pth{\frac{P^2}{n}+ P} = 0.
    \end{align*}
    If $m=n$, the solution is given by $\alpha^\ast = \beta^\ast = P/2$. If $m \neq n$, the solution is given by
    \begin{align}
        \alpha^\ast &= \sqrt{1+\frac{P}{n}} \pth{\frac{\sqrt{1+\frac{P}{m}}-\sqrt{1+\frac{P}{n}}}{\frac{1}{m}-\frac{1}{n}}}, \label{eq:optimal_alpha}\\
        \beta^\ast &= P -\alpha^\ast. \nonumber
    \end{align}
It is easy to verify that $(\alpha^\ast, \beta^\ast)$ is the unique minimizer to $f_P$ since it's convex over $S_P$. Now it remains to show the upper bound for $f_P(\alpha^\ast, \beta^\ast)$. Without loss of generality, in the reminder of the proof we assume $m < n$ and denote $\alpha^\ast$ by simply $\alpha$. Rewriting the optimal $\alpha$ in \prettyref{eq:optimal_alpha} in terms of $\snr= P/mn$, we obtain
\begin{align}
    \frac{\alpha}{mn} = \frac{\sqrt{(1+ n \, \snr)(1+ m \, \snr)} - (1+ m \, \snr)}{n-m}.
    \label{eq:alpha_rewri}
\end{align}
Now substituting this $\alpha$ and corresponding $\beta$ in $f_P(\alpha, \beta) =  \pth{1 + \frac{m}{\alpha}}
\pth{1 + \frac{n}{\beta}} $ and rearranging the terms, we get
\begin{align*}
    f_P(\alpha, \beta) &= 1 + \frac{1}{\snr} \, \pth{\frac{n-m}{mn}} \, \pth{\frac{1}{1- \frac{2 \alpha}{mn \, \snr}}} \\
    & = 1 + \frac{1}{n\, \snr} \pth{\frac{\frac{n}{m}-1}{1- \frac{2 \alpha}{mn \, \snr}}}. 
\end{align*}
Let $ \gamma \define \frac{m}{n} <1$. Now we study the behavior of $\alpha$ in \prettyref{eq:alpha_rewri} as a function of $\gamma$. In particular, define $g(\gamma) \define \sqrt{1+ n \, \snr} \, \sqrt{1+ n \gamma \, \snr} $. Observe that $g(1) = 1+ n \, \snr$ and $g'(1) = \frac{n \, \snr}{2}$. Rewriting \prettyref{eq:alpha_rewri} as a function of $\gamma$, we get
\begin{align*}
        \frac{\alpha}{mn} &= \frac{ g(\gamma) - (1+ n \gamma \, \snr)}{n(1-\gamma)} \\
        &= \frac{ g(1) +g'(1)(\gamma-1) - (1+ n \gamma \, \snr) + \frac{g''}{2}(\gamma-1)^2 + \frac{g'''}{3!}(\gamma-1)^3 + \ldots }{n(1-\gamma)} \\
        &= \frac{\snr}{2} + \frac{1}{n} \pth{\frac{g''}{2}(1-\gamma) - \frac{g'''}{3!}(1-\gamma)^2 + \ldots }.
\end{align*}
Utilizing the fact that $g''(1) = \frac{-1}{4}\, \frac{n^2 \snr^2}{1+ n \snr}$, $g''(1) = \frac{3}{8}\, \frac{n^3 \snr^3}{(1+ n \snr)^2}$ and so forth, we obtain
\begin{align*}
   1- \frac{2 \alpha}{mn \, \snr} &= \frac{2(1-\gamma)}{n \, \snr} \pth{\frac{1}{2}\, \frac{1}{4}\, \frac{n^2 \, \snr^2}{1+ n \, \snr} + \frac{1}{3!} \, \frac{3}{8}\, \frac{n^3 \, \snr^3}{(1+ n \,\snr)^2} (1-\gamma) + \ldots } \\
   & \geq \frac{2(1-\gamma)}{n \, \snr} \frac{1}{2}\, \frac{1}{4}\, \frac{n^2\, \snr^2}{1+ n \, \snr} \\
   & = \frac{(1-\gamma)}{4} \frac{n \,\snr}{1+ n \,\snr}.
\end{align*}
Substituting this bound back in the experssion for $f_P$ yields the final bound:
\begin{align*}
    f_P(\alpha, \beta) & \leq 1 + \frac{4}{n\gamma \, \snr} \pth{1+ \frac{1}{n \,\snr}} \\
    & = 1 + \frac{4}{m \, \snr} \pth{1+ \frac{1}{n \,\snr}}.
\end{align*}
\end{proof}

\subsection{Proof of \prettyref{lmm:power_rankr}}

\begin{proof}
    To minimize $f_P(\balpha, \bbeta)$ over $S_{P} =  \sth{ \{(\balpha, \bbeta) : \balpha \geq 0, \bbeta \geq 0, \sum_i (\alpha_i + \beta_i) = P }$, we consider a slightly relaxed version that serves as an upper bound to this problem. In particular, first we divide the power $P$ into $P_1,\ldots, P_r$ such that $\sum_i P_i = P$ and $P_i \geq 0$. Then for each $P_i$ we find the optimal $\alpha_i$ and $\beta_i$ from rank-$1$ allocation scheme in \prettyref{lmm:power_rankone} and compute the corresponding objective value. In the end, we find a tractable scheme for division of power $P$ among $P_1, \ldots, P_r$ minimizing this objective. Mathematically,
    \begin{align*}
        \min_{(\balpha, \bbeta) \in S_P} f_P(\balpha, \bbeta) & \leq \min_{\{\sum_i P_i = P\}} \min_{\{(\alpha_i, \beta_i): \alpha_i + \beta_i = P_i, i \in [r]\}} \sum_i \kappa_i \pth{1+ \frac{m} {\alpha_i}} \pth{1+ \frac{n}{\beta_i}} \\
        & = \min_{\{\sum_i P_i = P\}}  \sum_i \kappa_i \min_{(\alpha_i, \beta_i): \alpha_i + \beta_i = P_i} \pth{1+ \frac{m} {\alpha_i}} \pth{1+ \frac{n}{\beta_i}} \\
        & \stackrel{(\prettyref{lmm:power_rankone})}{\leq} \min_{\{\sum_i P_i = P\}}  \sum_i \kappa_i \pth{1 + \frac{4}{m \snr_i}\pth{1+ \frac{1}{n \snr_i}}}, \quad \snr_i \define \frac{P_i}{mn}, \\
        & = \min_{\{\sum_i P_i = P\}} \pth{ 1 + \frac{4}{m} \sum_i \frac{\kappa_i}{\snr_i} + \frac{4}{mn} \sum_i \frac{\kappa_i}{\snr_i^2} }.
    \end{align*}
Choosing $\snr_i \propto \sqrt{\kappa_i}$, \ie $\snr_i = \snr \, \frac{\sqrt{\kappa_i}}{\sum_j \sqrt{\kappa_j}}$, and substituting this allocation above, we obtain
\begin{align*}
    \min_{(\balpha, \bbeta) \in S_P} f_P(\balpha, \bbeta) & \leq  1+ \frac{4}{m \, \snr} \pth{\sum_i \sqrt{\kappa_i}}^2 + \frac{4}{mn \, \snr^2} \, R \pth{\sum_i \sqrt{\kappa_i}}^2 \\
    & \leq 1+ \frac{4}{(m/r) \, \snr}\pth{1 + \frac{4}{(n/r) \, \snr}},
\end{align*}
where we used the inequality $\pth{\sum_i \sqrt{\kappa_i}}^2 \leq r$ together with the fact that $\sum_i \kappa_i =1 $.
\end{proof}

\subsection{Proof of \prettyref{lmm:sgd_influ}}

\begin{proof}
Recall from ~\ref{eq:awgn_sgd} that the stochastic gradient reconstructed at the receiver after transmitting $\bg$ is $\by_{\gsgd}(\bg) \define \awgnp(\bg) = \bg + \frac{\norm{\bg}}{\sqrt{P}}\, \bZ$, where $\bZ$ is a zero-mean independent channel noise in $\reals^{m \times n}$. Thus
\begin{align*}
    \lambda_\gsgd = \frac{1}{\norm{\bg}^2} \, \Expect_{\bZ} \norm{\by_{\gsgd}(\bg) - \bg}^2 = \frac{1}{\norm{\bg}^2} \, \frac{\norm{\bg}^2}{P} \, \Expect \norm{\bZ}^2 = \frac{mn}{P} = \frac{1}{\snr}.
\end{align*}
\end{proof}

\subsection{Proof of \prettyref{lmm:miss_influ}}
\label{app:miss_influ}

\begin{proof}
In view of ~\ref{eq:laser_rewri}, denote the error compensated gradient at time $t$ as $\bM = \be_t + \gamma_t \bg_t$ and its compression as $\bM_r = \rankcx(\bM) = \sum_{i=1}^r \bp_i \bq_i^\top$ with orthogonal factors $\{\bp_i\}$ and orthonormal $\{\bq_i\}$ (without loss of generality). After transmitting these factors of $\bM_r$ via the noisy channel, we obtain 
\begin{align*}
\by_{\miss}(\bM_r) = \awgn_{(\balpha, \bbeta)}(\bM_r)  = \sum_{i=1}^r \pth{\bp_i + \frac{\norm{\bp_i}}{\sqrt{\alpha_i}} \cdot \bZ_m^{(i)}} \pth{\bq_i + \frac{\norm{\bq_i}}{\sqrt{\beta_i}} \cdot \bZ_n^{(i)}}^\top.
\end{align*}
Denote $\tilde{\bp}_i \define \bp_i + \frac{\norm{\bp_i}}{\sqrt{\alpha_i}} \cdot \bZ_m^{(i)}$, $\tilde{\bq}_i \define \bq_i + \frac{\norm{\bq_i}}{\sqrt{\beta_i}} \cdot \bZ_n^{(i)}$, and $\bZ = ( \bZ_m^{(i)},  \bZ_n^{(i)})_{i=1}^r$. We observe that $\Expect_{\bZ}[\by_{\miss}(\bM_r)] = \bM_r$. Hence

\begin{align*}
    \Expect_{\bZ} \norm{\by_{\miss}(\bM_r) - \bM_r}^2 &= \Expect_{\bZ}\norm{\sum_i \tilde{\bp}_i \tilde{\bq}_i^\top}^2 - \norm{\bM_r}^2 \\
    & = \sum_i \Expect_{\bZ} \norm{\tilde{\bp}_i}^2 \, \Expect_{\bZ} \norm{\tilde{\bq}_i}^2 - \sum_i \norm{\bp}^2 \norm{\bq}^2 \\
    & = \sum_i \norm{\bp}^2 \norm{\bq}^2 \qth{\pth{1+ \frac{m}{\alpha_i}} \pth{1+ \frac{n}{\beta_i}} - 1} \\
    & = \norm{\bM_r}^2 \pth{\sum_i \kappa_i \pth{1+ \frac{m}{\alpha_i}} \pth{1+ \frac{n}{\beta_i} } - 1} \\
    & \stackrel{(\small{\prettyref{lmm:power_rankr}})}{=} \norm{\bM_r}^2 \pth{f_P(\balpha, \bbeta) -1},
\end{align*}
where we set $\kappa_i = \norm{\bp_i}^2 /\norm{\bM_r}^2$. Now choosing $(\balpha, \bbeta) = (\balpha^\ast, \bbeta^\ast)$ as in \prettyref{lmm:power_rankr} yields the desired result. 
\end{proof}

\subsection{Proof of \prettyref{lmm:onestep_miss}}
\begin{proof}
    From \prettyref{eq:virtualrelative_miss}, we have that
    \begin{align*}
         \tilde{\btheta}_{t+1} = \tilde{\btheta}_t -\gamma_t \bg_t + \calC_r(\be_t + \gamma_t \bg_t) - \awgn_{(\balpha, \bbeta)}\pth{\calC_r(\be_t + \gamma_t \bg_t)}.
    \end{align*}
Denoting $\rm Error_{\bZ} = \calC_r(\be_t + \gamma_t \bg_t) - \awgn_{(\balpha, \bbeta)}\pth{\calC_r(\be_t + \gamma_t \bg_t)}$, we observe that $\Expect_{\bZ} [\rm Error_{\bZ}] = 0$ and $\Expect_{\bZ} \norm{\rm Error_{\bZ}}^2 \leq \missinflu \norm{\rankcx(\be_t + \gamma_t \bg_t)}^2 \leq  \missinflu \norm{\be_t + \gamma_t \bg_t}^2$ (see \prettyref{app:miss_influ}). Thus
\begin{align}
& \Expect \norm{\tilde{\btheta}_{t+1} - \btheta_\ast}^2 \nonumber \\
& = \Expect \norm{\tilde{\btheta}_{t} - \btheta_\ast - \gamma_t \bg_t}^2 + \Expect \norm{\rm Error_{\bZ}}^2 \nonumber  \\
& = \Expect \norm{\tilde{\btheta}_{t} - \btheta_\ast }^2 - 2 \gamma_t \Expect \inner{\bg_t}{\tilde{\btheta}_{t} - \btheta_\ast} + \gamma_t^2 \Expect \norm{\bg_t}^2 + \Expect \norm{\rm Error_{\bZ}}^2 \nonumber  \\
& \leq \Expect \norm{\tilde{\btheta}_{t} - \btheta_\ast }^2 - 2 \gamma_t \Expect \inner{\bg_t}{{\btheta}_{t} - \btheta_\ast} + 2 \gamma_t \Expect \inner{\bg_t}{{\btheta}_{t} - \tilde{\btheta}_{t}} + \gamma_t^2 \Expect \norm{\bg_t}^2 + \missinflu \Expect \norm{\be_t + \gamma_t \bg_t}^2 \nonumber  \\
& = \Expect \norm{\tilde{\btheta}_{t} - \btheta_\ast }^2 - 2 \gamma_t \Expect \inner{\bg_t}{{\btheta}_{t} - \btheta_\ast} + 2 \gamma_t \Expect \inner{\bg_t}{{\btheta}_{t} - \tilde{\btheta}_{t}} (1+\missinflu) + \gamma_t^2 \Expect \norm{\bg_t}^2 (1+\missinflu) \nonumber  \\
& \hspace{30em} +\missinflu\Expect \norm{\be_t}^2 \nonumber  \\
& \stackrel{(Assump.~\ref{assump:grad_oracle})}{\leq} \Expect \norm{\tilde{\btheta}_{t} - \btheta_\ast }^2 - 2 \gamma_t \Expect \inner{\fullgt}{{\btheta}_{t} - \btheta_\ast} + 2 \gamma_t \Expect \inner{\fullgt}{{\btheta}_{t} - \tilde{\btheta}_{t}} (1+\missinflu) \nonumber  \\
& \hspace{3em} +  (M+1) (1+\missinflu) \gamma_t^2  \Expect \norm{\fullgt}^2  
 + \gamma_t^2 \sigma^2 (1+\missinflu) + \missinflu \Expect \norm{\be_t}^2. \label{eq:quasi_help}
\end{align}
Now we closely follow the steps as in the proof of \cite{stich2020error}, Lemma 8. Since $f$ is $L$-smooth, we have $\norm{\fullgt}^2 \leq 2L(\fungap{t}$. Further, by Assumption~\ref{assump:quasi},
\begin{align*}
  -2\inner{\nabla f(\btheta_t)}{\btheta_t-\btheta_\ast} {\leq} - \mu \norm{\btheta_t- \btheta_\ast}^2  - 2(f(\btheta_t)-f_\ast)\,,
\end{align*}
and since $2\inner{\ba}{\bb} \leq \alpha \norm{\ba}^2 + \alpha^{-1} \norm{\bb}^2$ for $\alpha > 0$, $\ba,\bb \in \reals^d$, we have
\begin{align*}
2\inner{\nabla f(\btheta_t)}{\tilde \btheta_t - \btheta_t} \leq \frac{1}{2L} \norm{\nabla f(\btheta_t)}^2 + 2L\norm{\btheta_t -\tilde \btheta_t}^2 {\leq} f(\btheta_t)-f_\ast + 2L \norm{\btheta_t -\tilde \btheta_t}^2 \,. 
\end{align*}
And by $\norm{\ba + \bb}^2 \leq (1+\beta)\norm{\ba}^2 + (1+\beta^{-1}) \norm{\bb}^2$ for $\beta > 0$ (via Jensen's inequality), we observe
\begin{align*}
 -\norm{\btheta_t - \btheta_\ast}^2 \leq - \frac{1}{2} \norm{\tilde \btheta_t - \btheta_\ast}^2 + \norm{\btheta_t - \tilde \btheta_t}^2\,.
\end{align*}
Plugging these inequalities in \prettyref{eq:quasi_help}, we obtain that
\begin{align*}
    & \Expect \norm{\tilde{\btheta}_{t+1} - \btheta_\ast}^2 \\
    & \leq \pth{1- \frac{\mu \gamma_t}{2}}\Expect \norm{\tilde{\btheta}_{t} - \btheta_\ast}^2  - \gamma_t \pth{1- \missinflu -2L(M+1)(1+\missinflu)\gamma_t}\Expect(\fungap{t}) \\
    & \hspace{10em }+ \gamma_t^2 \sigma^2 (1+\missinflu) + (\mu \gamma_t + 2L\gamma_t (1+\missinflu)) \Expect \norm{\be_t}^2.
\end{align*}
Utilizing the fact that $\gamma_t \leq \frac{1- 2\lambda_\miss}{4L(M+1)(1+\missinflu)}$ and $\mu \leq L$ yields the desired claim.
\end{proof}

\subsection{Proof of \prettyref{lmm:misserror_control}}
\begin{proof}
    The proof of \prettyref{lmm:misserror_control} is very similar to that of \prettyref{lmm:quasi_errorbound} for ~\ref{eq:ef_sgd}. In that proof, a key step is to establish that $(3L(2/\delta + M)\gamma_t^2) \leq \frac{\delta}{64 L}$ and  $(3L \gamma_t\, 4/\delta) \leq 1$. In our setting, $\gamma_{t} \leq \frac{1}{10 L (2/\delta_r+M)(1+\lambda_\miss)}$ and $\missinflu \leq \frac{1}{10(2/\delta_r +M)}$. Thus
    \begin{align*}
        & \pth{3L(1+\lambda_\miss)+\frac{\lambda_\miss}{\gamma_t}} \gamma_t ^2 \pth{\frac{2}{\delta_r} + M} \\
        &= 3L\pth{\frac{2}{\delta_r} + M}(1+\lambda_\miss) \gamma_t \cdot \gamma_t + \lambda_\miss \pth{\frac{2}{\delta_r} + M} \gamma_t \\
        & \leq \frac{3}{10} \cdot \gamma_t + \frac{1}{10} \cdot \gamma_t \\
        & = \frac{4}{10}\, \frac{1}{10 L (\frac{2}{\delta_r}+M)(1+\lambda_\miss)} \\
        & \leq \frac{\delta_r}{32 L}.
    \end{align*}
Similarly,
\begin{align*}
     \frac{4}{\delta_r}\pth{3L(1+\lambda_\miss)\gamma_t +\lambda_\miss} &=  3L(1+\lambda_\miss) \, \frac{4}{\delta_r} \, \gamma_t + \missinflu \, \frac{4}{\delta_r} \\
     & \leq \frac{6}{10} + \frac{2}{10} \\
     & \leq 1.
\end{align*}
\end{proof}

\subsection{Proof of \prettyref{lmm:missnoncvx_progress}}
\begin{proof}
    From \prettyref{eq:virtualrelative_miss}, we have that
    \begin{align*}
         \tilde{\btheta}_{t+1} = \tilde{\btheta}_t -\gamma_t \bg_t + \calC_r(\be_t + \gamma_t \bg_t) - \awgn_{(\balpha, \bbeta)}\pth{\calC_r(\be_t + \gamma_t \bg_t)}.
    \end{align*}
Denoting $\rm Error_{\bZ} = \calC_r(\be_t + \gamma_t \bg_t) - \awgn_{(\balpha, \bbeta)}\pth{\calC_r(\be_t + \gamma_t \bg_t)}$, we observe that $\Expect_{\bZ} [\rm Error_{\bZ}] = 0$ and $\Expect_{\bZ} \norm{\rm Error_{\bZ}}^2 \leq \missinflu \norm{\rankcx(\be_t + \gamma_t \bg_t)}^2 \leq  \missinflu \norm{\be_t + \gamma_t \bg_t}^2$ (see \prettyref{app:miss_influ}). Using the smoothness of $f$,

\begin{align*}
    f(\tilde \btheta_{t+1}) \leq f(\tilde \btheta_{t}) - \gamma_t \inner{\nabla f(\tilde \btheta_{t})}{\bg_t} + \inner{f(\tilde \btheta_{t})}{\rm Error_{\bZ}} + \frac{L}{2} \norm{-\gamma_t \bg_t + \rm Error_{\bZ}}^2 
    \end{align*}
Taking expectation on both sides,
\begin{align*}
    \Expect f(\tilde \btheta_{t+1})
    & \leq \Expect f(\tilde \btheta_{t}) - \gamma_t \Expect \inner{\nabla f(\tilde \btheta_{t})}{\fullgt} + \frac{L}{2} \pth{\gamma_t^2 \Expect \norm{\bg_t}^2 + \missinflu \Expect \norm{\be_t + \gamma_t \bg_t}^2}.
\end{align*}

Rewriting $\inner{\nabla f(\tilde \btheta_{t})}{\fullgt} = \norm{\fullgt}^2 + \inner{\nabla f(\tilde \btheta_{t}) - \fullgt}{\fullgt} $ and using $\inner{\ba}{\bb} \leq \frac{1}{2}\norm{\ba}^2 + \frac{1}{2}\norm{\bb}^2$, we can simplify the expression as
\begin{align*}
    \inner{\nabla f(\tilde \btheta_{t}) - \fullgt}{\fullgt} &\leq \frac{1}{2}\norm{\nabla f(\btheta_t) - \nabla f(\tilde \btheta_t)}^2 + \frac{1}{2}\norm{\nabla f(\btheta_t)}^2\\
    & {\leq} \frac{L^2}{2}\norm{\btheta_t - \tilde \btheta_t}^2  + \frac{1}{2}\norm{\nabla f(\btheta_t)}^2\,.
\end{align*}
Pluggin this inequality back together with $\Expect\norm{\bg_t}^2 \leq (M+1) \Expect\norm{\fullgt}^2 + \sigma^2$, we get
\begin{align*}
    \Expect f(\tilde \btheta_{t+1})
    & \leq \Expect f(\tilde \btheta_{t}) - \frac{\gamma_t}{2}\pth{1- 2\gamma_t L(M+1)(1+\missinflu)} \Expect \norm{\fullgt}^2 + \frac{L \gamma_t^2 \sigma^2 (1+\missinflu)}{2} \\
    & \hspace{7em} + L \pth{\frac{L \gamma_t}{2} + \missinflu} \Expect \norm{\be_t}^2.
\end{align*}
Now utilizing the fact $\gamma_t \leq \frac{1}{4L(M+1)(1+\missinflu)}$ establishes the desired result.
\end{proof}

\section{Additional details about noisy channel and LASER}
\label{app:add_theory}

\subsection{Channel transformation}
\label{app:channel_transf}
Recall from \prettyref{eq:weightedgrad_awgn} in \prettyref{sec:background} that the server first obtains $\by = \sum_{i=1}^k a_i \bg_i + \bZ$, where $\norm{a_i \bg_i}^2 \leq P$ (note that we use the constant scheme $P_t=P$ as justified in \prettyref{sec:power-policy}). Now we want to show that for estimating the gradient sum $\sum_i \bg_i$ through a linear transformation on $\by$, the optimal power scalars are given by $a_i = \frac{\sqrt{P}}{\max_j \norm{\bg_j}}, \, \forall i \in [k]$, which yields the channel model in \eqref{eq:awgn_final}. 

Towards this, first let $k=2$ (the proof for general $k$ is similar). Thus our objective is
\begin{align*}
    \min_{a_1, a_2, b} \Expect \normbig{\frac{\by}{b} - \bg_1 - \bg_2}^2.
\end{align*}
For any $a_1, a_2, b$, we have that
\begin{align*}
\Expect \normbig{\frac{\by}{b} - \bg_1 - \bg_2}^2
    & =  \min_{a_1, a_2, b: \norm{a_i \bg_i}^2 \leq P} \Expect \normbig{\bg_1 \pth{\frac{a_1}{b}-1} + \bg_2 \pth{\frac{a_2}{b}-1} + \frac{\bZ}{b}}^2 \\
    & = \min_{a_1, a_2, b: \norm{a_i \bg_i}^2 \leq P} \Expect \normbig{\nabla f(\btheta)(\Delta_1 + \Delta_2)  + \Delta_1 \, \bxi_1 + \Delta_2 \, \bxi_2 + \frac{\bZ}{b}}^2, \quad \Delta_i = \frac{a_i}{b} -1 \\
    &= \min_{a_1, a_2, b: \norm{a_i \bg_i}^2 \leq P} \pth{\norm{\nabla f(\btheta)}^2 (\Delta_1 + \Delta_2)^2 + \Delta_1^2 \, \Expect \norm{\bxi_1}^2 + \Delta_2^2 \, \Expect \norm{\bxi_2}^2 + \frac{\Expect \norm{\bZ}^2}{b^2} },
\end{align*}
where we used the fact that $\bg_1 = \nabla f(\btheta) +\bxi_1$ and $\bg_2 = \nabla f(\btheta) +\bxi_2$ with zero-mean and independent $\bxi_1, \bxi_2$, and $\bZ$. We now observe that for any fixed $b$ the optimal $a_i$'s are given by $a_1 = a_2 = b$, \ie $\Delta_1 = \Delta_2 = 0$. To determine the optimal $b$, we have to solve
\begin{align*}
    \max \, b \quad \text{s.t.} \, \norm{b \, \bg_i}^2 \leq P,
\end{align*}
which yields $b^\ast = \sqrt{P}/{\max_i \norm{\bg_i}}$. The proof for general $k$ is similar.

\subsection{Detailed steps for \prettyref{algo:miss_comm}}
\label{app:detailed_algo}
Recall from \prettyref{algo:miss_comm} that power allocation among clients is done via the function $\textsc{poweralloc}( \{ \rankcx(\bM_j) , \bM_j \})$. The theoretically optimal power allocation is discussed in \prettyref{app:power_alloc}, and given explicitly in \prettyref{lmm:power_rankr}. However we empirically observe that we can relax this allocation scheme and even simpler schemes suffice to beat the other considered baselines. This is detailed in \prettyref{app:power-allocation-exp}.
\clearpage

\subsection{Constant-order SNR}
\label{app:constant_snr}

As discussed in \prettyref{sec:theory} and established in Lemmas ~ \ref{lmm:sgd_influ} and \ref{lmm:miss_influ} of \prettyref{app:channel_influ}, we have that

\begin{align*}
    \lambda_\miss \leq  \frac{4}{(m/r) \, \snr} \pth{1+ \frac{1}{(n/r) \, \snr}} \ll \frac{1}{\snr} = \lambda_{\gsgd}.
\end{align*}
In the low-rank \cite{vogels2019powersgd} and constant-order SNR regime where $r = \bigo(1)$ and $\snr = \Omega(1)$, we observe that $\lambda_\miss$ is roughly $\bigo(m)$ times smaller than $\lambda_{\gsgd}$. Note that this is only a sufficient theoretical condition to ensure that the ratio between $\lambda_\miss$ and $\lambda_{\gsgd}$ is smaller than one. In fact, a much weaker condition that $P/4r^2 > 1$ suffices. To establish this, we note
   $$
    \frac{ \lambda_\miss}{\lambda_{\gsgd}} = \frac{4r}{m} \pth{1+ \frac{r}{n \snr}} = \frac{4r}{m}\pth{1+ \frac{r m }{P}} = \frac{4r}{m}   + \frac{4r^2}{P}.
    $$

The first term is usually negligible since we always fix the rank $r=4$, which is much smaller compared to $m$ in the architectures we consider. Thus if $ P/4r^2 > 1 $, we see that the above ratio is smaller than one. Note that the constant-order SNR assumption already guarantees this: $\snr = \Omega (1) \Rightarrow P \gtrsim  m n \Rightarrow P \gtrsim r^2$, since $r$ is smaller than both $m$ and $n$. On the other hand, for the \resnet architecture with $L=61$ layers and $r=4$, the power levels $P= 250, 500$ violate the above condition as $P/(L r^2) < 4$ (note that the budget $P$ here is for the entire network and hence replaced by $P/L$). But empirically we still observe the accuracy gains in this low-power regime (\prettyref{fig:cifar10} in the paper).

\section{Experimental details}
\label{app:add_exp}

We provide technical details for the experiments demonstrated in \prettyref{sec:main_results}.

\subsection{\wikileaks experimental setup}
This section concerns the experimental details used to obtain \prettyref{fig:llm} and \prettyref{tab:llm-power-ratio} in the main text. \prettyref{tab:wikitext-setup} collects the settings we adopted to run our code. \prettyref{tab:GPT2-parameters} describes the model architecture, with its parameters, their shape and their uncompressed size.
% Generated with rank_based_methods.py
\begin{table}[h]
\caption{Default experimental settings for the \gptwo model used to learn the \wikileaks task.}
\label{tab:wikitext-setup}
\vspace{1mm}
\small%
\newcolumntype{R}{>{\raggedleft\arraybackslash}X}
\begin{tabularx}{\linewidth}{lX}
    \toprule
    % Parameter & Default value \\
    % \midrule
    Dataset & \wikileaks \\
    Architecture & \gptwo (as implemented in \cite{pagliardini-llm}) \\
    \midrule
    Number of workers & 4 \\
    Batch size & $15$ per worker \\
    Accumulation steps & 3 \\
    \midrule
    Optimizer & AdamW ($\beta_1 = 0.9, \beta_2 = 0.95$) \\
    Learning rate & $0.001$ \\
    Scheduler & Cosine \\
    \# Iterations & $20000$ \\
    Weight decay & $\num{1e-3}$\\
    \midrule
    Dropout & $0.2$ \\
    Sequence length & $512$ \\
    Embeddings & $768$ \\
    Transformer layers & $12$ \\
    Attention heads & $12$ \\
    \midrule
    Power budget & 6 levels: 10k, 40k, 160k, 640k, 2560k, 10240k \\
    Power allocation & Proportional to norm of compressed gradients (uncompressed gradients for \textsc{Z-SGD})\\
    Compression & Rank 4 for \miss; $0.2$ compression factor for other baselines \\
    \midrule
    Repetitions & 1 \\
    \bottomrule
\end{tabularx}
\end{table}
% Generated with rank_based_methods.py
\begin{table}[h]
\caption{Parameters in the GPT-2 architecture, with their shape and uncompressed size.}
\label{tab:GPT2-parameters}
\vspace{1mm}
\small%
\newcolumntype{R}{>{\raggedleft\arraybackslash}X}
\begin{tabularx}{\linewidth}{Xllr}
\toprule
Parameter
& Gradient tensor shape
& Matrix shape
& Uncompressed size\\
\cmidrule(lr){1-4}
transformer.wte 
& $50304 \times 768$
& $50304 \times 768$
& 155 MB\\
transformer.wpe 
& $512 \times 768$
& $512 \times 768$
& 1573 KB\\
transformer.h.ln\_1 $(\times 12)$
& $768$
& $768 \times 1$
& $(12 \times)$ 3 KB\\
transformer.h.attn.c\_attn $(\times 12)$
& $2304 \times 768$
& $2304 \times 768$
& $(12 \times)$ 7078 KB\\
transformer.h.attn.c\_proj $(\times 12)$
& $768 \times 768$
& $768 \times 768$
& $(12 \times)$ 2359 KB\\
transformer.h.ln\_2 $(\times 12)$
& $768$
& $768 \times 1$
& $(12 \times)$ 3 KB\\
transformer.h.mlp.c\_fc $(\times 12)$
& $3072 \times 768$
& $3072 \times 768$
& $(12 \times)$ 9437 KB\\
transformer.h.mlp.c\_proj $(\times 12)$
& $768 \times 3072$
& $768 \times 3072$
& $(12 \times)$ 9437 KB\\
transformer.ln\_f
& $768$
& $768 \times 1$
& 3 KB\\
\cmidrule(lr){1-4}
\textbf{Total}
&
&
& 496 MB \\
\bottomrule
\end{tabularx}
\end{table}
\clearpage

\subsection{\cifarten experimental setup}
This section concerns the experimental details used to obtain \prettyref{fig:cifar10} and \prettyref{tab:power-ratio} in the main text. \prettyref{tab:cifar10-setup} collects the settings we adopted to run our code. \prettyref{tab:ResNet18-parameters} describes the model architecture, with its parameters, their shape and their uncompressed size.
% Generated with rank_based_methods.py
\begin{table}[h]
\caption{Default experimental settings for the \resnet model used to learn the \cifarten task.}
\label{tab:cifar10-setup}
\vspace{1mm}
\small%
\newcolumntype{R}{>{\raggedleft\arraybackslash}X}
\begin{tabularx}{\linewidth}{lX}
    \toprule
    % Parameter & Default value \\
    % \midrule
    Dataset & \cifarten \\
    Architecture & \resnet \\
    \midrule
    Number of workers & 16 \\
    Batch size & $128$ per worker \\
    \midrule
    Optimizer & SGD\\
    Momentum & 0.9 \\
    Learning rate & Grid-searched in $\{0.001, 0.005, 0.01, 0.05\}$ for each power level \\
    \# Epochs & 150 \\
    Weight decay & $\num{1e-4}$,\\
                  &$0$ for BatchNorm parameters \\
    \midrule
    Power budget & 10 levels: $250, 500, 1000, 2000, 4000, 8000, 16000, 32000, 64000, 128000$ \\
    Power allocation & Proportional to norm of compressed gradients (uncompressed gradients for \textsc{Z-SGD})\\
    Compression & Rank 4 for \miss; $0.2$ compression factor for other baselines \\
    \midrule
    Repetitions & 3, with varying seeds \\
    \bottomrule
\end{tabularx}
\end{table}
% Generated with rank_based_methods.py
\begin{table}[h]
\caption{Parameters in the ResNet18 architecture, with their shape and uncompressed size.}
\label{tab:ResNet18-parameters}
\vspace{1mm}
\small%
\newcolumntype{R}{>{\raggedleft\arraybackslash}X}
\begin{tabularx}{\linewidth}{Xllr}
\toprule
Parameter
& Gradient tensor shape
& Matrix shape
& Uncompressed size\\
\cmidrule(lr){1-4}
layer4.1.conv2 
& $512 \times 512 \times 3 \times 3$
& $512 \times 4608$
& 9437 KB\\
layer4.0.conv2 
& $512 \times 512 \times 3 \times 3$
& $512 \times 4608$
& 9437 KB\\
layer4.1.conv1 
& $512 \times 512 \times 3 \times 3$
& $512 \times 4608$
& 9437 KB \\
layer4.0.conv1 
& $512 \times 256 \times 3 \times 3$
& $512 \times 2304$
& 4719 KB \\
layer3.1.conv2 
& $256 \times 256 \times 3 \times 3$
& $256 \times 2304$
& 2359 KB \\
layer3.1.conv1 
& $256 \times 256 \times 3 \times 3$
& $256 \times 2304$
& 2359 KB \\
layer3.0.conv2 
& $256 \times 256 \times 3 \times 3$
& $256 \times 2304$
& 2359 KB \\
layer3.0.conv1 
& $256 \times 128 \times 3 \times 3$
& $256 \times 1152$
& 1180 KB \\
layer2.1.conv2 
& $128 \times 128 \times 3 \times 3$
& $128 \times 1152$
& 590 KB \\
layer2.1.conv1 
& $128 \times 128 \times 3 \times 3$
& $128 \times 1152$
& 590 KB \\
layer2.0.conv2 
& $128 \times 128 \times 3 \times 3$
& $128 \times 1152$
& 590 KB \\
layer4.0.shortcut.0 
& $512 \times 256 \times 1 \times 1$
& $512 \times 256$
& 524 KB \\
layer2.0.conv1 
& $128 \times 64 \times 3 \times 3$
& $128 \times 576$
& 295 KB \\
layer1.1.conv1 
& $64 \times 64 \times 3 \times 3$
& $64 \times 576$
& 147 KB \\
layer1.1.conv2 
& $64 \times 64 \times 3 \times 3$
& $64 \times 576$
& 147 KB \\
layer1.0.conv2 
& $64 \times 64 \times 3 \times 3$
& $64 \times 576$
& 147 KB \\
layer1.0.conv1 
& $64 \times 64 \times 3 \times 3$
& $64 \times 576$
& 147 KB \\
layer3.0.shortcut.0 
& $256 \times 128 \times 1 \times 1$
& $256 \times 128$
& 131 KB \\
layer2.0.shortcut.0 
& $128 \times 64 \times 1 \times 1$
& $128 \times 64$
& 33 KB \\
linear 
& $10 \times 512$
& $10 \times 512$
& 20 KB \\
conv1 
& $64 \times 3 \times 3 \times 3$
& $64 \times 27$
& 7 KB \\
Bias vectors (total)
&
&
& 38 KB \\
\cmidrule(lr){1-4}
\textbf{Total}
&
&
& 45 MB \\
\bottomrule
\end{tabularx}
\end{table}
%\raggedbottom
%\pagebreak
\clearpage

\subsection{\cifarhun experimental results}
This section concerns experimental results on \cifarhun. We used the same \resnet architecture as for \cifarten (except for the final layer, adapted to the 100-class dataset). We once again compared \miss to the usual baselines. \prettyref{fig:cifar100} and \prettyref{tab:cifar100-power-ratio} collect the results that we obtained. It can be seen that \miss outperforms the other algorithms with an even wider margin compared to the \cifarten and \wikileaks tasks, with a power gain of around $32 \times$ across different accuracy targets. \signum is much more sensitive to noise and performs much worse than the other algorithms; therefore, we decided to leave out its results in order to improve the quality of the plot. \prettyref{tab:cifar100-setup} collects the settings we adopted to run our code. \prettyref{tab:ResNet18-parameters-100} describes the model architecture, with its parameters, their shape and their uncompressed size.
% Generated with rank_based_methods.py
\begin{table}[h]
\caption{Default experimental settings for the \resnet model used to learn the \cifarhun task.}
\label{tab:cifar100-setup}
\vspace{1mm}
\small%
\newcolumntype{R}{>{\raggedleft\arraybackslash}X}
\begin{tabularx}{\linewidth}{lX}
    \toprule
    % Parameter & Default value \\
    % \midrule
    Dataset & \cifarhun \\
    Architecture & \resnet \\
    \midrule
    Number of workers & 16 \\
    Batch size & $128$ per worker \\
    \midrule
    Optimizer & SGD\\
    Momentum & 0.9 \\
    Learning rate & Grid-searched in $\{0.001, 0.005, 0.01, 0.05\}$ for each power level \\
    LR decay & $/10$ at epoch 150 \\
    \# Epochs & 200 \\
    Weight decay & $ \num{1e-4} $ \\
                  &$0$ for BatchNorm parameters \\
    \midrule
    Power budget & 10 levels: $500, 1000, 2000, 4000, 8000, 16000, 32000, 64000, 128000, 256000$ \\
    Power allocation & Proportional to norm of compressed gradients (uncompressed gradients for \textsc{Z-SGD})\\
    \midrule
    Repetitions & 3, with varying seeds \\
    Compression & Rank 4 for \miss; $0.2$ compression factor for other baselines \\
    \bottomrule
\end{tabularx}
\end{table}
% Generated with rank_based_methods.py
\begin{table}[h]
\caption{Parameters in the ResNet18 architecture, with their shape and uncompressed size.}
\label{tab:ResNet18-parameters-100}
\vspace{1mm}
\small%
\newcolumntype{R}{>{\raggedleft\arraybackslash}X}
\begin{tabularx}{\linewidth}{Xllr}
\toprule
Parameter
& Gradient tensor shape
& Matrix shape
& Uncompressed size\\
\cmidrule(lr){1-4}
layer4.1.conv2 
& $512 \times 512 \times 3 \times 3$
& $512 \times 4608$
& 9437 KB\\
layer4.0.conv2 
& $512 \times 512 \times 3 \times 3$
& $512 \times 4608$
& 9437 KB\\
layer4.1.conv1 
& $512 \times 512 \times 3 \times 3$
& $512 \times 4608$
& 9437 KB \\
layer4.0.conv1 
& $512 \times 256 \times 3 \times 3$
& $512 \times 2304$
& 4719 KB \\
layer3.1.conv2 
& $256 \times 256 \times 3 \times 3$
& $256 \times 2304$
& 2359 KB \\
layer3.1.conv1 
& $256 \times 256 \times 3 \times 3$
& $256 \times 2304$
& 2359 KB \\
layer3.0.conv2 
& $256 \times 256 \times 3 \times 3$
& $256 \times 2304$
& 2359 KB \\
layer3.0.conv1 
& $256 \times 128 \times 3 \times 3$
& $256 \times 1152$
& 1180 KB \\
layer2.1.conv2 
& $128 \times 128 \times 3 \times 3$
& $128 \times 1152$
& 590 KB \\
layer2.1.conv1 
& $128 \times 128 \times 3 \times 3$
& $128 \times 1152$
& 590 KB \\
layer2.0.conv2 
& $128 \times 128 \times 3 \times 3$
& $128 \times 1152$
& 590 KB \\
layer4.0.shortcut.0 
& $512 \times 256 \times 1 \times 1$
& $512 \times 256$
& 524 KB \\
layer2.0.conv1 
& $128 \times 64 \times 3 \times 3$
& $128 \times 576$
& 295 KB \\
layer1.1.conv1 
& $64 \times 64 \times 3 \times 3$
& $64 \times 576$
& 147 KB \\
layer1.1.conv2 
& $64 \times 64 \times 3 \times 3$
& $64 \times 576$
& 147 KB \\
layer1.0.conv2 
& $64 \times 64 \times 3 \times 3$
& $64 \times 576$
& 147 KB \\
layer1.0.conv1 
& $64 \times 64 \times 3 \times 3$
& $64 \times 576$
& 147 KB \\
layer3.0.shortcut.0 
& $256 \times 128 \times 1 \times 1$
& $256 \times 128$
& 131 KB \\
layer2.0.shortcut.0 
& $128 \times 64 \times 1 \times 1$
& $128 \times 64$
& 33 KB \\
linear 
& $100 \times 512$
& $100 \times 512$
& 205 KB \\
conv1
& $64 \times 3 \times 3 \times 3$
& $64 \times 27$
& 7 KB \\
Bias vectors (total)
&
&
& 38 KB \\
\cmidrule(lr){1-4}
\textbf{Total}
&
&
& 45 MB \\
\bottomrule
\end{tabularx}
\end{table}
\begin{table}
  \centering
  \small
  \begin{minipage}[t]{0.50\textwidth}
    \centering
    \includegraphics[width=\textwidth]{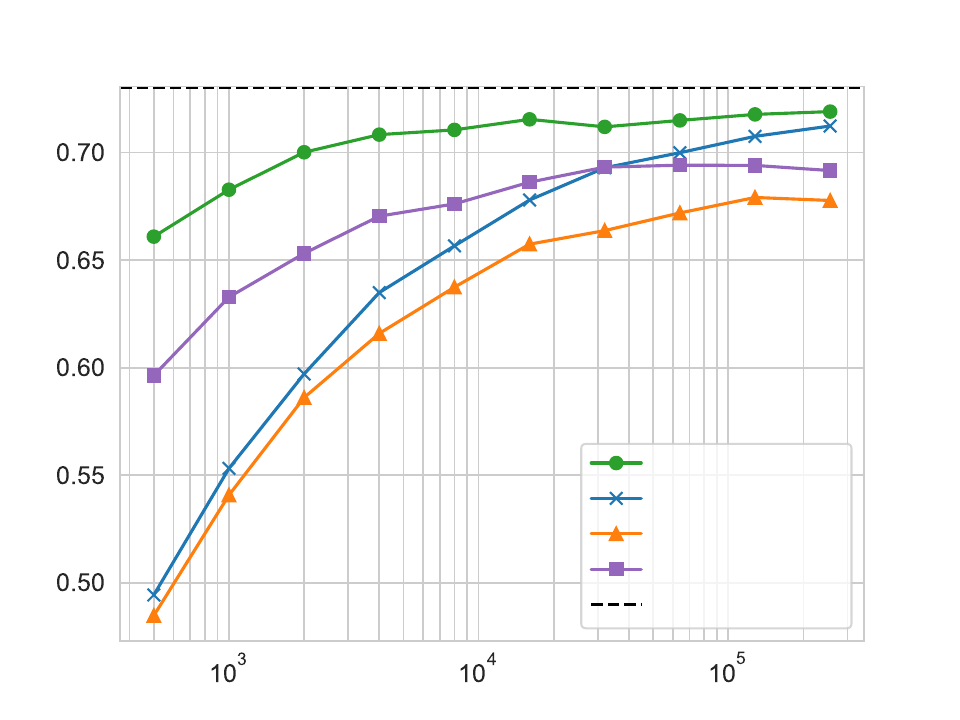}
      \put(-120,-1){\fontsize{8}{3}\selectfont Power budget}
      \put(-200,70){\rotatebox[origin=t]{90}{\fontsize{8}{3}\selectfont Accuracy}}
      \put(-61.5,51.5){\fontsize{6}{3}\selectfont LASER}
      \put(-61.5,44){\fontsize{6}{3}\selectfont Z-SGD}
      \put(-61.5,36.5){\fontsize{6}{3}\selectfont Sketching}
      \put(-61.5,29){\fontsize{6}{3}\selectfont Random-K}    
      \put(-61.5,22){\fontsize{6}{3}\selectfont Noiseless SGD}
      \vspace{5pt}
    \captionof{figure}{Test accuracy (\emph{higher the better}) for a given power budget on \textsc{Cifar-100} for different algorithms. The advantage of \miss is evident across the entire power spectrum.}
    \label{fig:cifar100}
  \end{minipage}
  \hspace{12pt}
  \begin{minipage}[b]{0.42\textwidth}
    \caption{Power required (\emph{lower the better}) to reach the given target accuracy on \textsc{Cifar-100}. \miss requires $16-32 \times $ lesser power than the \textsc{Z-SGD} to achieve the same targetaccuracy. Equivalently, \miss tolerates more channel noise than the \textsc{Z-SGD} for the same target accuracy as is partly supported by our theoretical analysis. }
    \vspace{10pt}
	\label{tab:cifar100-power-ratio}
	\begin{tabularx}{\textwidth}{c c c c}
    \toprule
    \multirow{2}{*}{\bf Target} &
    \multicolumn{2}{c}{\bf Power required} & 
    \multirow{2}{*}{\bf Reduction} \\
    
    \cmidrule(lr){2-3}
    
      & \miss & \textsc{Z-SGD} & \\
    
    \cmidrule(lr){1-1}\cmidrule(lr){2-3}\cmidrule(lr){4-4}
    
    $65\%$ & $500$ & $8000$ & $16\times$\\
    $68\%$ & $1000$ & $32000$ & $32\times$\\
    $70\%$ & $2000$ & $64000$ & $32\times$\\
    $71\%$ & $8000$ & $256000$ & $32\times$\\
    \bottomrule
    \end{tabularx}
  \end{minipage}
  \vspace{-3mm}
\end{table}
\clearpage

\subsection{\mnist experimental setup}
This section concerns the experimental details used to obtain \prettyref{tab:mnist} in the main text. \prettyref{tab:mnist-setup} collects the settings we adopted to run our code.
% Generated with rank_based_methods.py
\begin{table}[h]
\caption{Default experimental settings for the \textsc{1-Layer NN} used to learn the \mnist task.}
\label{tab:mnist-setup}
\vspace{1mm}
\small%
\newcolumntype{R}{>{\raggedleft\arraybackslash}X}
\begin{tabularx}{\linewidth}{lX}
    \toprule
    % Parameter & Default value \\
    % \midrule
    Dataset & \mnist \\
    Architecture & \textsc{1-Layer NN} \\
    \midrule
    Number of workers & 16 \\
    Batch size & $128$ per worker \\
    \midrule
    Optimizer & SGD\\
    Momentum & 0.9 \\
    Learning rate & $0.01$ \\
    \# Epochs & 50 \\
    Weight decay & $\num{1e-4}$,\\
    \midrule
    Power budget & 3 levels: $0.1, 1, 10$ \\
    Power allocation & Proportional to norm of compressed gradients (uncompressed gradients for \textsc{Z-SGD})\\
    \midrule
    Repetitions & 3, with varying seeds \\
    Compression & Rank 2 for \miss; $0.1$ compression factor for other baselines \\
    \bottomrule
\end{tabularx}
\end{table}
%\raggedbottom
%\pagebreak
%\clearpage

\subsection{Rank-accuracy tradeoff}
\label{sec:rank-accuracy}

\begin{figure}[h]
    \centering
    \includegraphics[width=0.6\textwidth]{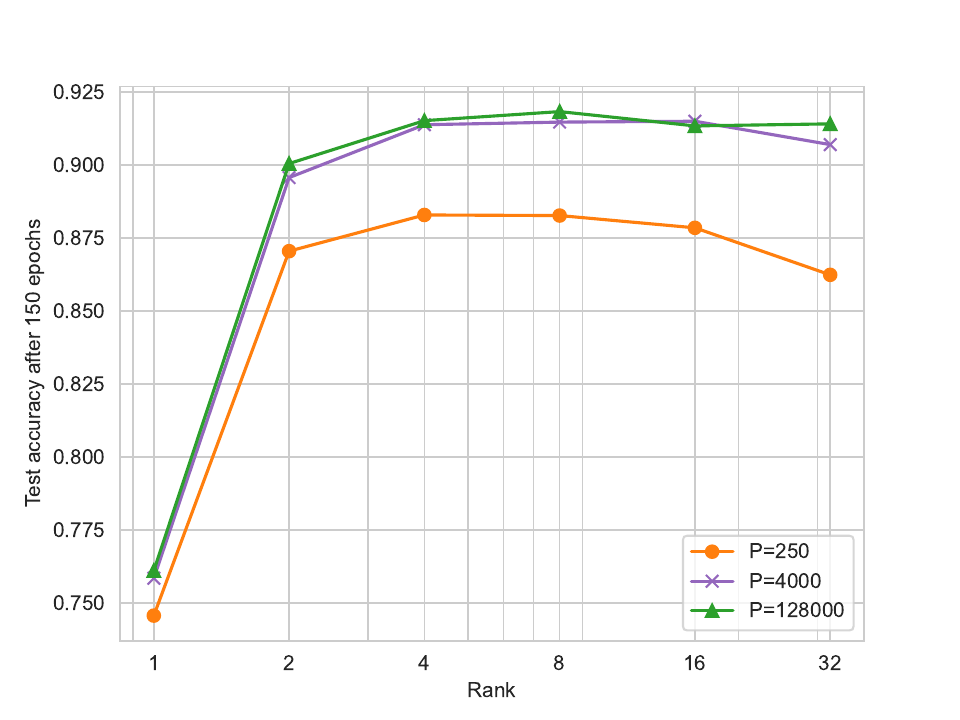}
    \caption{Final accuracy vs. compression rank tradeoff for CIFAR-10 classification, for low, medium and high power regimes. Rank-$4$/Rank-$8$ compression is optimal for all the three regimes. It reveals two interesting insights: (i) performance is uniformly worse in all the regimes with overly aggressive rank-one compression,  and (ii) higher rank compression impacts low power regime more significantly than the medium and high-power counterparts. This confirms with the intuition that at low power (and hence noisier channel), it is better to allocate the limited power budget appropriately to few ``essential'' rank components as opposed to thinning it out over many.}
    \label{fig:rank-accuracy}
\end{figure}

There exists an inherent tradeoff between the decomposition rank $r$ (and hence the compression factor $\delta_r$) and the final model accuracy. In fact, a small rank $r$ implies aggressive compression and hence the compression noise dominates the channel noise. Similarly, for a high decomposition rank, the channel noise overpowers the compression noise as the power available per each coordinate is small. We empirically investigate this phenomenon for \cifarten classification over various power regimes in \prettyref{fig:rank-accuracy}.

As \prettyref{fig:rank-accuracy} reveals, either Rank-$4$ or Rank-$8$ compression is optimal for all the three power regimes. Further we observe two interesting trends: (i) the final accuracy is uniformly worse in all the regimes with overly aggressive rank-one compression, and (ii) higher rank compression impacts the low power regime more significantly than the medium and high-power counterparts. This is in agreement with the intuition that at low power (and hence noisier channel), it is better to allocate the limited power budget appropriately to few “essential” rank components as opposed to thinning it out over many. This phenomenon can be theoretically explained by characterizing the compression factor $\delta_r$ as a function of rank $r$ and its effect on the model convergence. While the precise expression for $\delta_r$ is technically challenging, given the inherent difficulty in analyzing the PowerSGD algorithm \cite{vogels2019powersgd}, we believe that a tractable characterization of this quantity (via upper bounds etc.) can offer fruitful insights into the fundamental rank-accuracy tradeoff at play.

\begin{wrapfigure}{r}{0.49\textwidth}
%\begin{figure}[t]
  \begin{center}
  \vspace{-1.5em}
 \includegraphics[width=0.49\textwidth]{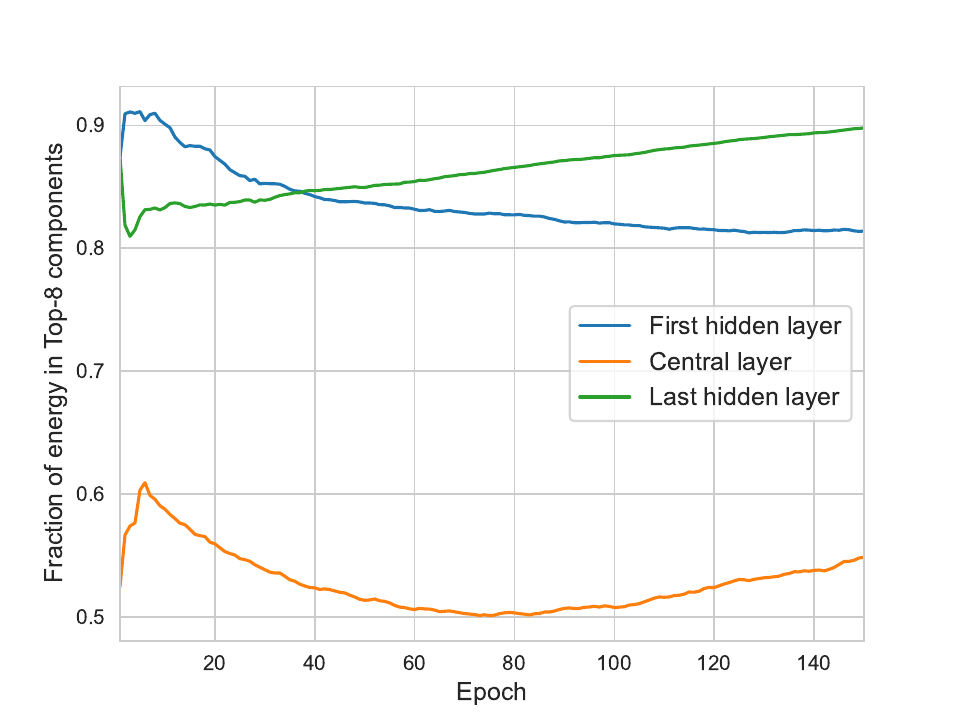}
 \end{center}  
\vspace{-1em}
     \caption{Fraction of energy in the top 8 components of the gradients of three layers in the network: the first and last hidden layer, and one central layer.}
    \label{fig:rank-study}
%\end{figure}
\vspace{-1em}
\end{wrapfigure}

To further shed light on this phenomenon, we trained the noiseless \ref{eq:sgd} on \cifarten and captured the evolution across the epochs of the energy contained in the top eight components of each gradient matrix. As illustrated in \prettyref{fig:rank-study}, we observe that for the first and last hidden layers, $80\%$ of the energy is already captured in these eight components. On the other hand, for the middle layer this fares around $55\%$. It is interesting to further explore this behavior for GPT models and other tasks.

\subsection{Power allocation across workers and neural network parameters}
\label{app:power-allocation-exp}
The choice of power allocation over the layers of the network is perhaps the most important optimization required in our experimental setup. Notice that, because of \prettyref{eq:weightedgrad_awgn}, all clients must allocate the same power to a given gradient, since otherwise it would be impossible to recover the correct average gradient. However, workers have a degree of freedom in choosing how to distribute the power budget among gradients, \ie among the layers of the network, and this power allocation can change over the iterations of the model training. 

\prettyref{app:power_alloc} analyzes power allocation optimality from a theoretical point of view. On the experimental side, simpler schemes are enough to get significant gains over the other baselines. As a matter of fact, we considered the following power allocation scheme for the experiments: at each iteration, each worker determines locally how to allocate its power budget across the gradients. Then, we assume that this power allocation choice is communicated by the client to the server noiselessly. The server then takes the average of the power allocation choices, and communicates the final power allocation to the clients. The clients then use this power allocation to send the gradients to the server via the noisy channel.

For the determination of each worker's power allocation, three schemes were considered: 
\begin{itemize}
    \item uniform power to each gradient;
    \item power proportional to the Frobenius norm (or the square of it) of the gradients;
    \item power proportional to the norm of the compressed gradients (i.e., the norm of what is actually communicated to the server).
\end{itemize}
For \textsc{Z-SGD}, where there is no gradient compression, the best power allocation turned out to be the one proportional to the norm of the gradients, independently of the power constraint imposed. For all the other algorithms, the best is power proportional to the norm of the compressed gradients.

\subsection{Static vs. dynamic power policy}
\label{app:power-policy}

\begin{figure}[h]
\centering
\includegraphics[width=0.7\textwidth]{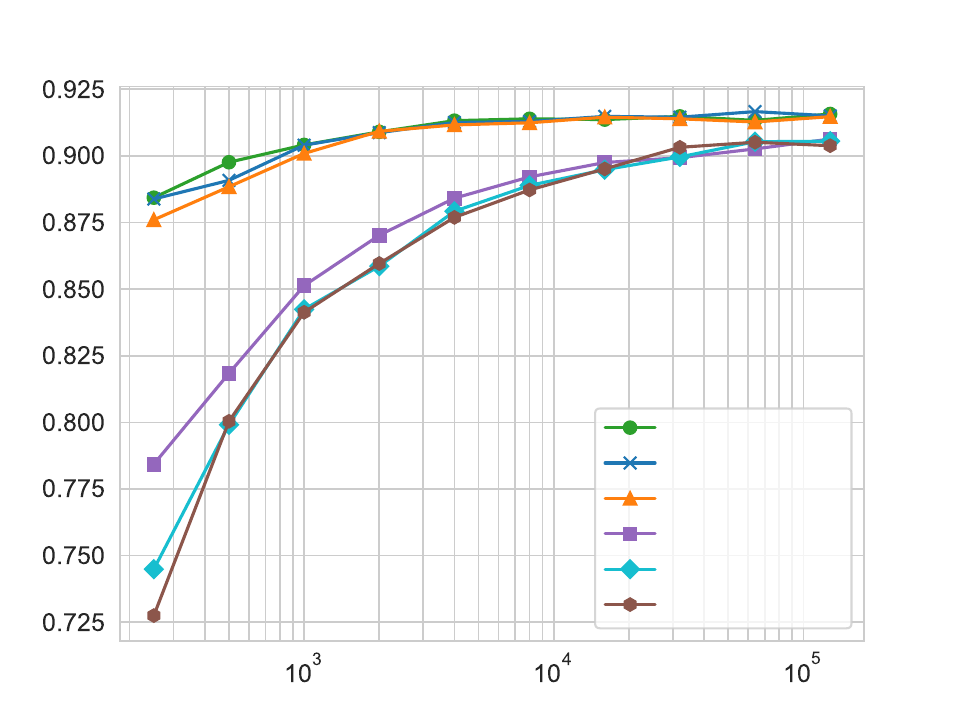}
    \put(-163,-1){\fontsize{9}{3}\selectfont Power budget}
    \put(-280,95){\rotatebox[origin=t]{90}{\fontsize{9}{3}\selectfont Accuracy after 150 epochs}}
    \put(-84,82.5){\fontsize{7}{3}\selectfont LASER, constant}
    \put(-84,72.5){\fontsize{7}{3}\selectfont LASER, linear}
    \put(-84,62){\fontsize{7}{3}\selectfont LASER, step}
    \put(-84,52){\fontsize{7}{3}\selectfont Z-SGD, constant}    
    \put(-84,41.5){\fontsize{7}{3}\selectfont Z-SGD, linear}
    \put(-84,31.5){\fontsize{7}{3}\selectfont Z-SGD, step}
\caption{Final accuracy vs. power budget $P$ with various power control schemes, for distributed training across $16$ workers with \resnet on \cifarten. For each budget $P$, we consider three increasing power control laws, as studied in the literature [1], that satisfy the average power constraint: (i) constant power, $P_t = P$, (ii) piecewise constant, with the power levels $P_t \in \{P/3, 2P/3, P, 4P/3, 5P/3\}$, and (iii) linear law between the levels $P/3$ and $5P/3$. The performance of increasing power allocation schemes is equal or worse compared to their decreasing counterparts of \prettyref{fig:power-policy}.}
\label{fig:power-policy-increasing}
\end{figure}
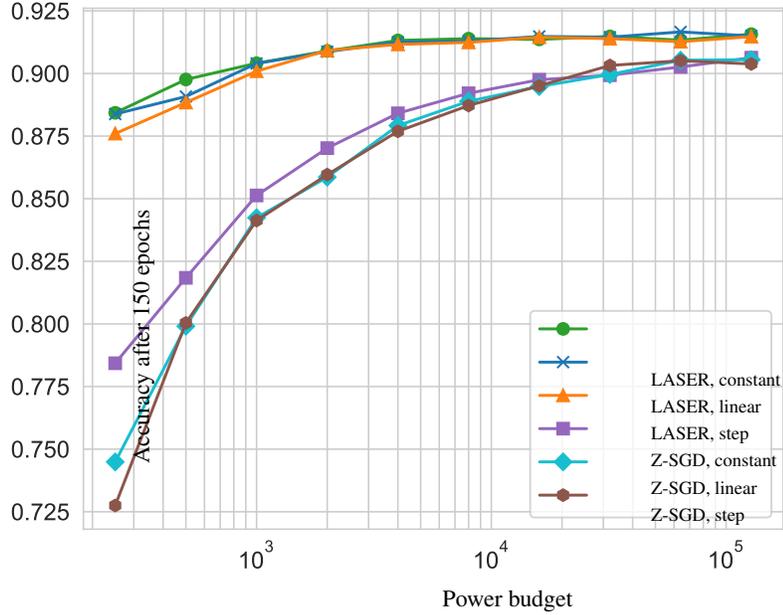

As discussed in \prettyref{sec:power-policy}, we analyzed different power allocation schemes across iterations, when a fixed budget in terms of average power over the epochs is given. \prettyref{fig:power-policy} shows the results for decreasing power allocations, while \prettyref{fig:power-policy-increasing} here shows their increasing counterparts. We observe that LASER exhibits similar gains over Z-SGD for all the power control laws. Further, constant power remains the best policy for both LASER and Z-SGD. Whilst matching the constant power performance, the power-decreasing control performs better than the increasing counterpart for Z-SGD, especially in the low-power regime, where the accuracy gains are roughly $4-5\%$.

\clearpage

\subsection{Baselines implementation}
In this section we describe our implementation of the baselines considered in the paper.

\subsubsection{Count-Mean Sketching}
\begin{algorithm}[H]
  \caption{\textsc{Count-Mean Sketching}}\label{alg:sketching}
  \begin{algorithmic}[1]
  \Function{compress}{gradient matrix $M \in \mathbb{R}^{n \times m}$}
    \State Treat $M$ as a vector of length $nm$.
    \State The number of samples $b$ is set to $mn \times \text{(compression factor)}$.
    \State If the resulting $b$ is less than 1, we set $b=1$.
    \State Sample a set of $mn$ indices $I$ i.i.d. between $0$ and $b-1$ using the same seed on all workers.
    \State Sample a set of $mn$ signs ($+1$ or $-1$) $S$ i.i.d. using the same seed used for $I$.
    \State $\hat C \leftarrow \0 \in \mathbb{R}^{b}$
    \For{$j=0,\dots,mn-1$}
    \State $\hat C (I(j)) \leftarrow \hat C (I(j)) + S(j) \times M(j)$
    \EndFor
    \State \Return $\hat C$
  \EndFunction
  \Function{aggregate+decompress}{worker's values $\hat{C}_1\ldots \hat{C}_k$}
    \State Sample $I$ and $S$ as before, using the same seed.
    \State $\hat M \leftarrow \0 \in \mathbb{R}^{n \times m}$
    \State $\hat M(I) \leftarrow \frac{1}{k} \sum_{i=1}^k \hat{C}_i(I) \odot S$
    \State \Return $\hat{M}$
  \EndFunction
  \end{algorithmic}
\end{algorithm}
Power is allocated proportional to compressed gradients' norms. The algorithm is implemented without local error feedback, since error feedback causes the algorithm to diverge. The compression factor was grid-searched in $\{0.1, 0.2, 0.5, 0.8\}$ and $0.2$ was finally chosen as the overall best.

\subsubsection{Random K}
\begin{algorithm}[H]
  \caption{Random $K$}\label{alg:random_k}
  \begin{algorithmic}[1]
  \Function{compress}{gradient matrix $M \in \mathbb{R}^{n \times m}$}
    \State Treat $M$ as a vector of length $nm$.
    \State The number of samples $b$ is set to $mn \times \text{(compression factor)}$.
    \State If the resulting $b$ is less than 1, we set $b=1$.
    \State Sample a set of $b$ indices $I$ without replacement, using the same seed on all workers.
    \State \Return Looked up values $S=M(I)$.
  \EndFunction
  \Function{aggregate+decompress}{worker's values $S_1\ldots S_k$}
    \State $\hat M \leftarrow \0 \in \mathbb{R}^{n \times m}$
    \State $\hat M(I) \leftarrow \frac{1}{k} \sum_{i=1}^k S_i$
    \State \Return $\hat M$
  \EndFunction
  \end{algorithmic}
\end{algorithm}
Power is allocated proportional to compressed gradients' norms. The algorithm is implemented with local error feedback. The compression factor was grid-searched in $\{0.1, 0.2, 0.5, 0.8\}$ and $0.2$ was finally chosen as the overall best.

\subsubsection{Signum}
\begin{algorithm}[H]
  \caption{\signum}\label{alg:signum}
  \begin{algorithmic}[1]
  \Function{compress}{gradient matrix $M \in \mathbb{R}^{n \times m}$}
    \State Compute the signs $S \in \left\{-1,1\right\}^{n \times m}$ of $M$
    \State \Return $S$
  \EndFunction
  \Function{aggregate+decompress}{worker's signs $S_1 \ldots S_k$}
    \State \Return $\textsc{sign}(\sum_{i=1}^k S_i)$
  \EndFunction
  \end{algorithmic}
\end{algorithm}
We implemented \signum following \cite{bernstein2018signsgd}. We run it in its original form, without error feedback. Power is allocated proporional to the compressed gradients' norms. Since the compressed gradients are simply the sign matrices, in this case power is allocated proportional to the square root of the number of parameters in each layer $\sqrt{mn}$. Unlike the other baselines, \signum does not require any compression factor.

\end{document}